\documentclass{article}
\usepackage{tabularx}

\PassOptionsToPackage{super,sort&compress,comma}{natbib}

\usepackage[preprint]{neurips_2024}

\usepackage[utf8]{inputenc} %
\usepackage[T1]{fontenc}    %
\usepackage{hyperref}       %
\usepackage{url}            %
\usepackage{booktabs}       %
\usepackage{amsfonts}       %
\usepackage{amsmath,amssymb}        %
\usepackage{nicefrac}       %
\usepackage{microtype}      %
\usepackage{xcolor}         %
\usepackage{graphicx}
\usepackage{todonotes}
\usepackage{array}
\usepackage{wrapfig}
\usepackage{afterpage}
\usepackage{placeins}
\usepackage{lipsum}
\usepackage{multirow}
\usepackage{longtable}
\usepackage[version=4]{mhchem}
\usepackage{siunitx}
\usepackage{algorithm}
\usepackage{algorithmic}
\usepackage{tabularx}
\usepackage{booktabs}
\title{Large language models as uncertainty-calibrated optimizers for experimental discovery}

\author{
  Bojana Ranković\textsuperscript{1,2}, \quad  Ryan-Rhys Griffiths\textsuperscript{3} \quad Philippe Schwaller\textsuperscript{1,2} \\
  \textsuperscript{1}\'Ecole Polytechnique F\'{e}d\'{e}rale de Lausanne (EPFL) \\
  \textsuperscript{2}National Centre of Competence in Research (NCCR) Catalysis \\
  \textsuperscript{3}Independent Researcher \\
  \texttt{\{bojana.rankovic,philippe.schwaller\}@epfl.ch} \\
}

\begin{document}

\maketitle

\begin{abstract}

    Scientific discovery increasingly depends on efficient experimental optimization to navigate vast design spaces under time and resource constraints. Traditional approaches often require extensive domain expertise and feature engineering. While large language models, with their vast scientific knowledge, circumvent the feature engineering limitations, they lack the calibrated uncertainty estimates required for high-stakes decision making. Hence, current optimization methods force a choice between domain knowledge and reliability, with no principled approach that affords both. In this work, we show that training language models through the uncertainty-aware objectives of traditional optimization methods enables their use as reliable optimizers guided by natural language. By teaching LLMs from experimental outcomes under uncertainty, we transform their overconfidence from a fundamental limitation into a precise calibration mechanism. Applied to Buchwald-Hartwig reactions, a cornerstone of pharmaceutical synthesis, our method nearly doubles the discovery rate of high-yielding reaction conditions -- 24\% to 43\% in 50 experimental iterations starting from 10 unsuccessful conditions. Across 19 diverse optimization problems spanning organic synthesis, materials science and catalysis, process chemistry, and molecular design, our approach ranks first on average, establishing a new paradigm for reliable, uncertainty-guided optimization with LLMs. Our approach can accelerate discovery by lowering the barrier to using powerful optimization methods, replacing the need for domain-specific feature engineering with more accessible natural language interfaces. These findings highlight that ensuring reliability through principled uncertainty quantification is critical for realizing the full potential of AI-guided experimentation.

\end{abstract}

\section{Main}

Artificial intelligence can now generate millions of novel drug candidates, material structures, or chemical reactions in a matter of hours, expanding an already vast combinatorial landscape that has long challenged experimental science \cite{butler2018machine, 2018_Segler, stokes2020deep}. However, the fundamental bottleneck remains: deciding which candidates to evaluate next. This explosion of possibilities demands systematic, sample-efficient approaches to navigate vast design spaces under time and resource constraints \cite{mervin2021uncertainty, tavazza2021uncertainty, 2021_Griffiths, schwaller2022machine}.

Established experimental optimization methods, such as Bayesian optimization (BO) \cite{kushner1962versatile, kushner1964new, garnett2023bayesian, schweidtmann2018machine, muller2022automated, torres2022multi, hickman2022equipping}, address this problem through a probabilistic lens. BO relies on surrogate models, typically Gaussian processes (GPs) \citep{2006_Williams}, to provide experimental predictions together with uncertainties, balancing exploration and exploitation \cite{jones1998efficient, jones2001taxonomy, auer2002using}. In practice, however, BO typically depends on domain-specific feature representations \cite{tran2018active, shields2021bayesian, pomberger2022effect, rankovic2023bayesian}. Optimizing a chemical reaction, for example, requires first translating molecular structures, catalytic mechanisms, and reaction pathways into descriptors that encode chemical intuition \cite{taylor2023brief, guo2023bayesian}. When shifting domains from organic synthesis to materials discovery, the time-consuming feature engineering must be repeated, as descriptors for one field do not apply to the other. Reaction fingerprints cannot characterize crystal structures, catalyst optimization bears little resemblance to drug design, and even within catalysis, descriptors optimized for a metal-ligand system may not be effectively transferred to others \cite{durand2019computational, van2022physics, zaza2025holistic}. Although BO provides a general approach, each new domain effectively starts from scratch with little performance guarantee, creating barriers to adoption and limiting the transferability of expertise \cite{bai2023transfer}.

In contrast, large language models (LLMs) have emerged as generalist models that transcend domain boundaries through rich prior knowledge acquired from their vast pretraining data \citep{vaswani2017attention, devlin2018bert, radford2018improving, wei2022emergent}. The potential use of LLMs for experimental optimization has attracted considerable interest, yet adoption remains constrained by a fundamental flaw: unreliability. LLMs hallucinate, producing outputs that sound confident but are factually incorrect \citep{maynez2020faithfulness, peng2023check, huang2025survey}. Although this limitation frustrates users in everyday applications, it has severe consequences in experimental science. Hallucinated experimental suggestions waste resources, delay research timelines, and compromise safety protocols in automated laboratory settings \cite{musslick2025automating, ramos2025review}. Unsurprisingly, their integration into high-stakes experimental workflows remains impractical without rigorous control \cite{bran2023chemcrow, zhang2025exploring}.

The merging of the two paradigms, BO and LLMs, has therefore gained significant traction \citep{jablonka202314, ramos2023bayesian, rankovic2023bochemian, kristiadi2024sober, agarwalsearching, cisse2025language}. Yet existing approaches suffer because they treat language models and probabilistic optimization separately. We argue instead that the resolution lies in recognizing they should exist as a unified system. Here, we introduce GOLLuM (Gaussian Process Optimized LLMs) as the first framework to train language models through the same probabilistic objective that makes Bayesian optimization reliable. Rather than positioning LLMs as direct experimental optimizers, where they struggle with hallucination and heuristic search, we demonstrate how their probabilistic integration within established optimization mechanisms enables principled decision-making from natural language.

The framework establishes a joint optimization mechanism between LLM embeddings and probabilistic modeling. Given textual descriptions of prior experiments -- reaction conditions, synthesis procedures, process parameters -- the LLM transforms these into embeddings serving as inputs to a GP surrogate. Critically, the learning signals flow back from the GP to update the LLM. Through this bidirectional mechanism, the LLM learns to encode experimental performance rather than textual similarity, organizing its latent space into distinct regions of high- and low-performing experiments.

This learned organization reveals structure in the experimental landscape while providing interpretability. In Buchwald-Hartwig cross-coupling reactions, fundamental to medicinal chemistry, the transformed representations expose genuine chemical patterns explaining reaction performance. The model automatically groups reactions by their key chemical components, clustering iodide-based reagents in high-yield regions while separating them from low-performing bromides and chlorides. This separation mirrors known reactivity trends, offering rationale for why recommended conditions should work -- yet it emerges entirely from the experimental data, without explicit chemical instruction. By exploiting these discovered patterns, GOLLuM nearly doubles the discovery rate of high-performing conditions, reaching 43\% of the highest ranking conditions within 50 iterations compared to 24\% with static embeddings. This improvement extends to other reaction classes, such as Suzuki-Miyaura cross couplings and decarboxylative arylations, while still outperforming domain-specialized descriptors. Most remarkably, the approach transfers across domains: from materials synthesis and process engineering to molecular design. Across 19 diverse benchmarks covering different optimization paradigms (combinatorial, mixed variables, continuous), GOLLuM ranks first on average using fixed hyperparameters (tuned once on a single dataset) starting from 10 failed experiments (below the median).

Our results suggest that uncertainty, rather than being LLMs' fatal flaw for experimental science, can drive their transformation into reliable optimizers. By training language models together with the GP surrogates, we enable what neither paradigm could achieve alone: the generality to handle any experimental procedures described in natural language with the reliability required for real experimental campaigns. This principle of uncertainty-driven adaptation suggests a broader paradigm for deploying AI in high-stakes domains where calibrated confidence matters as much as capability.

\begin{figure}[h!]
\begin{center}
\includegraphics[width=\linewidth]{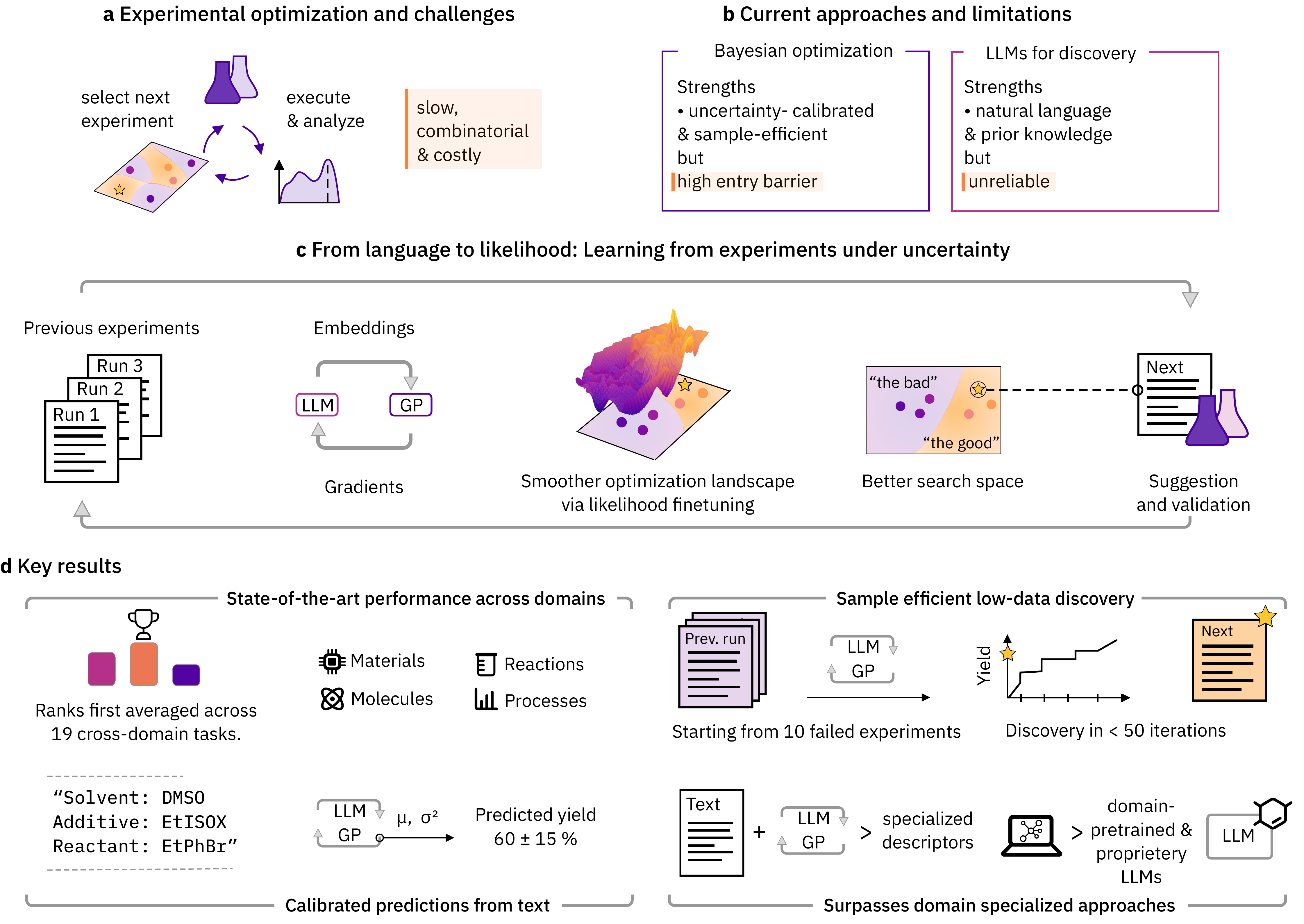}
\end{center}
\caption{\textbf{Language models as uncertainty-calibrated optimizers} \textbf{a}, Experimental optimization challenges. Scientists face the universal problem of efficiently exploring vast design spaces under time and cost constraints. \textbf{b}, Current approaches impose a trade-off between accessibility and reliability. Bayesian optimization provides principled uncertainty calibration and sample efficiency but requires domain expertise. LLMs offer natural language interfaces and prior knowledge but lack reliability.
\textbf{c}, GOLLuM optimization loop. Previous experimental results, described in natural language, feed into joint optimization of both GP and LLM components. Likelihood gradients from the GP finetune the LLM, progressively smoothening the optimization landscape and organizing the search space into distinct regions: "the good", "the bad" (and the ugly) of high and low performance. This structured representation enables easier selection for the next experiment, which becomes a new training point that continues the loop. \textbf{d}, Key results. GOLLuM ranks first averaged across 19 cross-domain tasks spanning materials, reactions, molecules, and processes, enabling sample-efficient discovery from natural language inputs while surpassing both specialized descriptors, domain-pretrained models, and proprietary LLMs.}
\label{fig:overview}
\end{figure}

\section{Results}

\subsection{Large language models are unreliable as direct optimizers}

Using large language models as direct, prompt-based optimizers is an appealing strategy for leveraging their vast prior knowledge in scientific discovery \cite{yang2023large, ramos2023bayesian, nguyen2024predicting, macknight2025pre}. In such scenarios, the LLM directly proposes experimental conditions based on the prompt containing details of the design space, previous experiments, and the optimization goal. However, this application relies on the model's black-box patterns and ignores known limitations, including overconfidence and a lack of principled uncertainty calibration. 

\begin{figure}[h!]
\begin{center}
\includegraphics[width=0.99\linewidth]{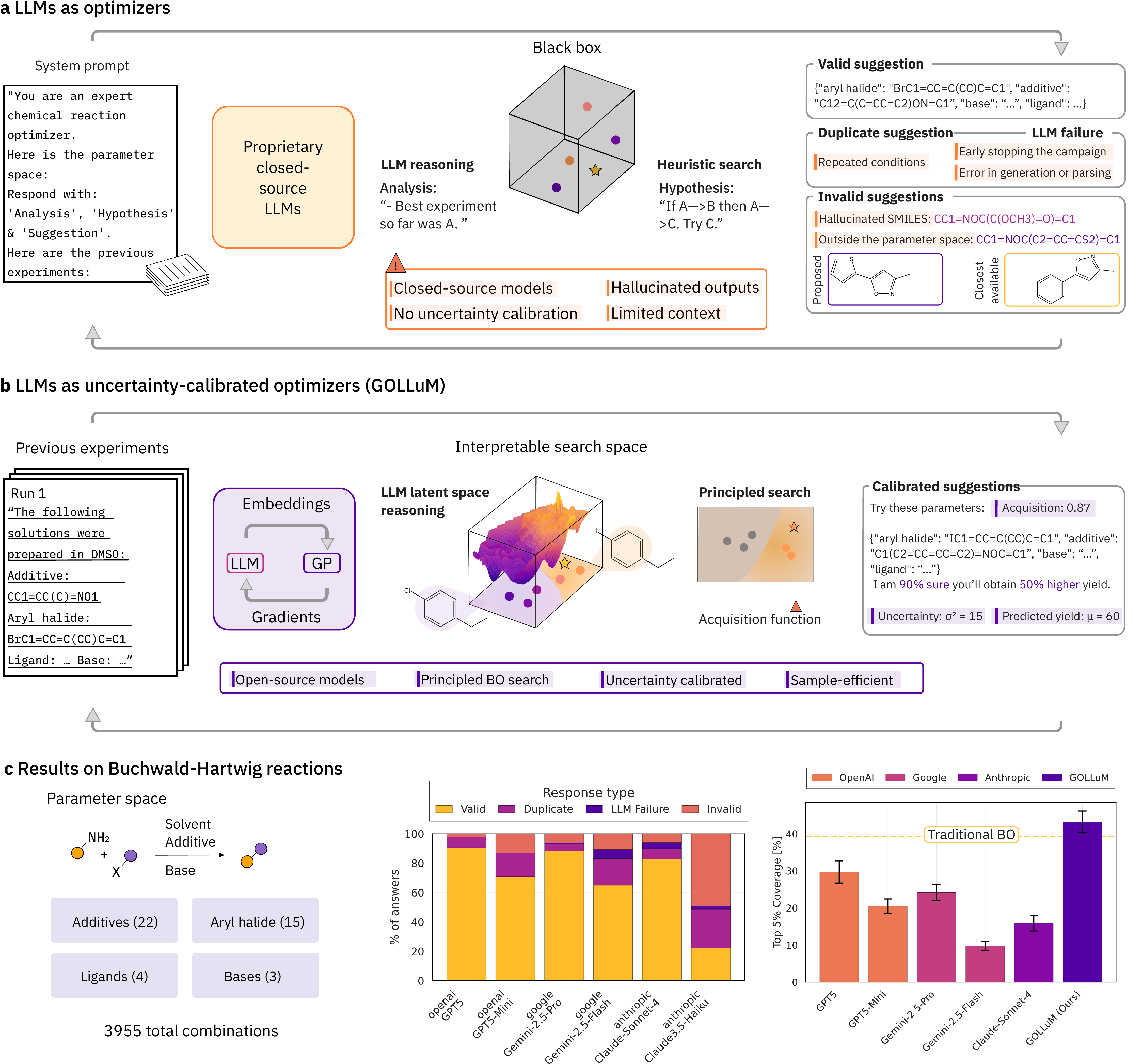}
\end{center}
\caption{\textbf{From LLM prompting to LLMs as principled scientific optimizers.} \textbf{a}, Querying LLMs for optimization. Given the parameter space and previous experiments, the LLM proposes next suggestions. This method relies on heuristic search and suffers from a lack of uncertainty calibration, which leads to invalid suggestions, hallucinations, and frequent failures. \textbf{b}, GOLLuM framework integrates an LLM and a GP in a joint optimization loop. Gradients from the GP's probabilistic objective finetune the LLM, creating an interpretable search space that enables principled, uncertainty-aware, and sample-efficient optimization. \textbf{c}, Quantitative results on the Buchwald-Hartwig reactions benchmark. (Left) The combinatorial design space of 3955 reactions across 5 distinct chemical products, corresponding to $\sim$800 unique conditions per product. (Middle) The analysis of response types from several proprietary LLMs reveals failure rates, with a high fraction of invalid or duplicated suggestions. We executed 5 independent optimization seeds (API-cost limited) per product. Each run cold-started with 10 initial low-performing reactions and repeatedly queried the model until reaching a quota of 50 unique, in-space suggestions. The stacked bars aggregate the raw outputs from all queries issued while reaching that quota across 25 runs (5 products x 5 seeds) per model ($>$1250 queries). Categories: Valid, Duplicate (already proposed in the same run), Invalid (outside the parameter space), and LLM failure (format/parse/timeout). (Right) A comparison of optimization performance ("Top 5\% Coverage") shows that direct LLM approaches substantially underperform traditional Bayesian optimization (with reaction fingerprints \cite{probst2022reaction}) and our proposed GOLLuM method using GP-finetuned general-purpose open-source LLMs (T5 \cite{2020t5}). Bars show mean and standard error. Performance comparison excludes Claude-3.5 Haiku (did not reach the quota of 50 valid in-space suggestions).} 
\label{fig:llms}
\end{figure}

We first evaluated this approach for experimental optimization by prompting LLMs to act as expert chemical optimizers (Figure \ref{fig:llms}a). We analyzed results of several state-of-the-art proprietary, closed-source LLMs to suggest new conditions for Buchwald-Hartwig \cite{ahneman2018predicting} reactions, providing them with the parameter space and a history of previous experiments. This approach, however, proved fundamentally unreliable for scientific discovery.

A quantitative analysis of the models' responses revealed a high failure rate across all tested LLMs (10\% to $\sim$80\% in Figure \ref{fig:llms}c, Response type). A significant fraction of suggestions were invalid due to hallucinated chemical structures or proposals outside the defined parameter space. Furthermore, models frequently suggested duplicate conditions, failed to generate parsable outputs or tried to prematurely end the optimization campaign. These failures stem from the inherent limitations of using LLMs as direct optimizers: their lack of principled uncertainty calibration and their reliance on heuristic search rather than a principled probabilistic strategy. 

While LLMs offer integrated reasoning and contextual analysis of the optimization problem, there are currently no clear mechanisms to evaluate or mitigate their overconfidence or outright hallucinations in a quantifiable manner. This black-box reasoning process makes them unsuitable for high-stakes experimental decision-making and underscores the need for a framework that integrates their domain knowledge within a reliable, uncertainty-aware optimization loop. Critically, when benchmarked against traditional Bayesian optimization, the LLM-driven approaches substantially underperformed, failing to consistently identify the top-performing reactions, and trailing behind our proposed GOLLuM method (Figure \ref{fig:llms}c, Top 5\% Coverage).

The failure of direct prompting motivates a more principled approach. Besides their generative abilities, LLMs offer their scientific knowledge in a more constrained manner: as sophisticated feature extractors for a traditional Bayesian optimization framework. We next investigate this strategy by using static LLM embeddings as chemical descriptors.

\subsection{Static LLM embeddings underperform as chemical descriptors}

A critical requirement for scientific optimization is a representation that captures the relationship between experimental parameters and outcomes \cite{rankovic2023bayesian}. Traditional chemical descriptors rely on structured data, leaving the vast majority of scientific knowledge constrained in unstructured language such as patents, publications, and even chemical intuition. Here, we evaluate whether LLMs can translate the chemical knowledge from text into data-driven descriptors for Bayesian optimization tasks. 

\begin{figure}[h!]
\begin{center}
\includegraphics[width=0.99\linewidth]{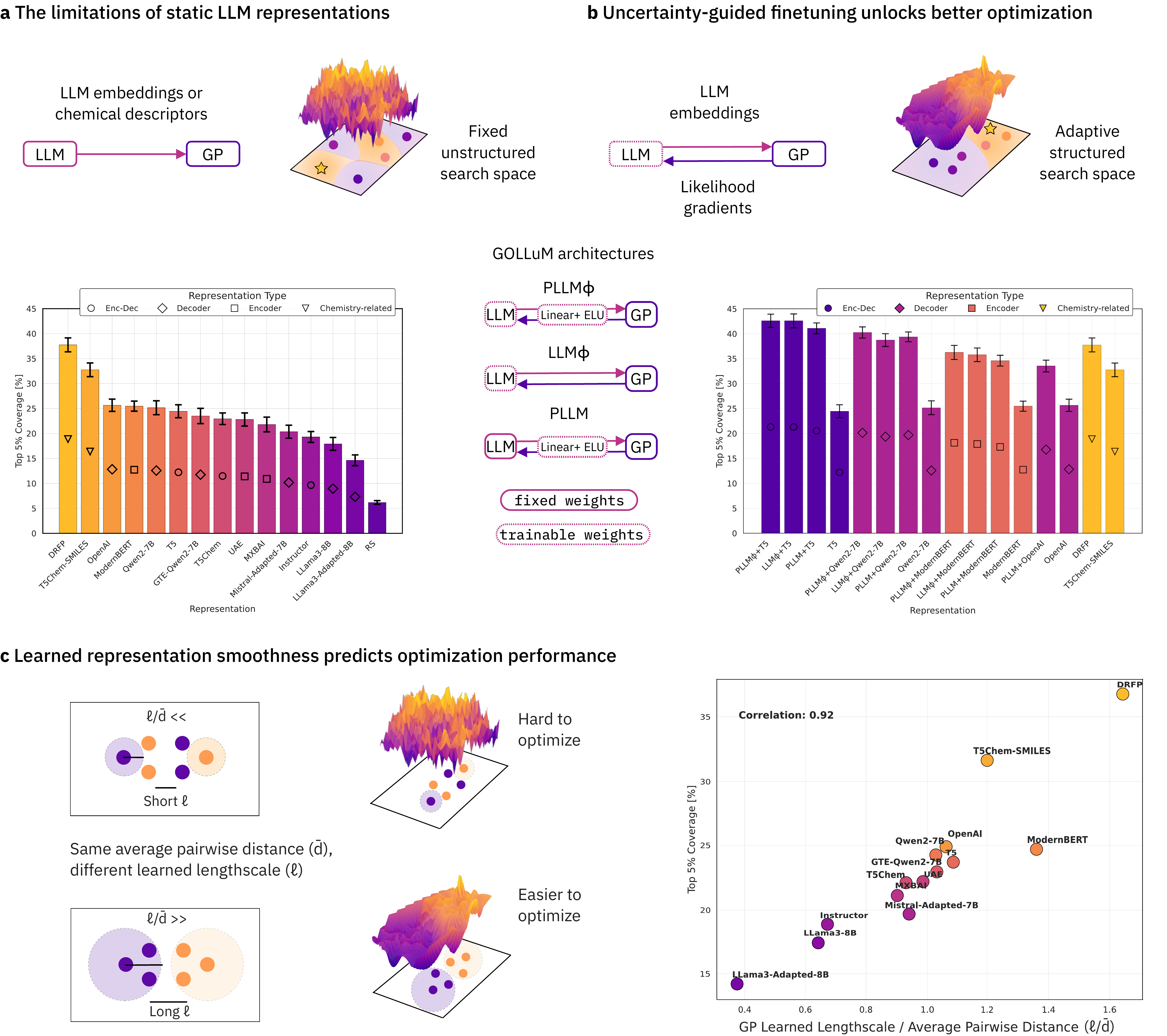}
\end{center}
\caption{\textbf{Representation geometry enables efficient Bayesian optimization} \textbf{a}, BO performance with fixed LLM features as input to GP. We benchmark three encoder models (ModernBERT \cite{modernbert}, UAE \cite{li2023angle}, MXBAI \cite{emb2024mxbai}), three encoder--decoders (Instructor \cite{su2022one}, T5 \cite{2020t5} and its chemistry-related variant T5Chem \cite{christofidellis2023unifying}) and four decoder families (Llama \cite{grattafiori2024llama, behnamghader2024llm2vec}, Mistral \cite{behnamghader2024llm2vec}, Qwen \cite{bai2023qwen, li2023towards}, and OpenAI \cite{openai2024new}). We show average discovery of high-impact regions of the design space as the percentage of the top 5\% reactions found during the optimization, aggregated across all five Buchwald-Hartwig reactions. Chemistry-related representations include T5Chem-SMILES\cite{christofidellis2023unifying}, a pretrained chemistry-related LLM with SMILES input, and DFRP\cite{probst2022reaction}, a reaction fingerprint. \textbf{b}, Comparative analysis of GP-based LLM finetuning. The finetuned models are arranged by the overall performance and relative improvements to their base (fixed embeddings) LLM-GP variants. Chemistry-related baselines (previous best) included for comparison. All results show mean and standard error across 20 independent seeds, with each optimization run starting from 10 initial points (random below median) and selecting subsequent points via acquisition function maximization over the enumerated design space. \textbf{c}, Data representations and their success rates in BO. BO performance correlates with GP smoothness, measured as the ratio of learned lengthscale to average pairwise embedding distance. Higher ratios indicate smoother GP fits, implying the representation supports meaningful generalization across the design space.} 
\label{fig:bh-results}
\end{figure}

On the same benchmark set of Buchwald–Hartwig reactions \cite{ahneman2018predicting}, we observed that static LLM representations consistently underperformed specialized chemical descriptors (Figure \ref{fig:bh-results}a). In 50 optimization trials initiated from a realistic cold-start scenario with 10 low-performing experiments, LLM embeddings helped discover only 15–26\% of the highest-yielding reactions (5th percentile threshold), trailing behind the 38\% achieved by specialized chemical fingerprints (DRFP \cite{probst2022reaction}). OpenAI’s embeddings \cite{openai2024new}, while reaching the highest score among the LLM embeddings, still failed to close this performance gap.

The simple lack of chemical knowledge does not explain these differences. Chemistry-specialized language models, such as T5Chem \cite{christofidellis2023unifying}, underperformed general-purpose variants, achieving a 23\% discovery rate despite extensive pretraining on chemical tasks. The model's dependence on input format revealed underlying patterns that explain the disparity in results. Changing T5Chem's input from an experimental procedure to the Simplified Molecular Input Line Entry System (SMILES) format \cite{weininger1990smiles, anderson1987smiles, weininger1988smiles} improved its discovery rate by 44\%. This improvement is not surprising given the model's inherent familiarity with the SMILES notation from its pretraining. Such format dependency indicates the core issue is not the knowledge the model contains, but rather its ability to translate that knowledge into the embedding space.

These results highlight the limitations of static LLM embeddings for experimental guidance. Without direct adaptation to the optimization task, they force a trade-off between the convenience of natural language and the optimization performance of structured descriptors.

\subsection{Representation geometry determines optimization success}
\label{sec:smoothness-ratio}

The representation geometry determines optimization success better than predictive accuracy of a surrogate model (Figure \ref{fig:bh-results}c). Although regression metrics such as $R^2$ explain how well the surrogate fits the objective function, they do not offer a clear answer as to what dictates the optimization performance. Instead, we identify a simple metric based on representation geometry that correlates better with optimization success.

The ratio of the GP’s learned lengthscale ($\ell$) to the average pairwise distance between points in the search space ($\overline{d}$) robustly predicts optimization success. The lengthscale defines the distance in the input space over which the model expects output values to be correlated. A higher $\ell/\overline{d}$ indicates a smoother, more GP-friendly embedding space where the model can generalize over larger area because the embeddings position similar outcome experiments together. Across all fourteen representations, this geometric measure correlates strongly with BO performance ($r = 0.92$, versus $R^2: r = 0.78$, weighted $R^2: r = 0.82$, see Figure \ref{fig:struc-vs-fit-corr} in SI). By contrast, representations with a low smoothness ratio create rough, difficult-to-model objective landscapes with potentially many local minima.

These findings underscore a well-structured representation as a more important factor for successful optimization than the raw predictive accuracy itself. If a GP surrogate with a stationary kernel (where the relationship between two points depends only on their distance) assumes a smooth objective function, the underlying representation should support it. Specialized chemical descriptors (DRFP \cite{probst2022reaction}) instill some organization in the latent space, resulting in a higher ratio ($\ell/\overline{d} \approx 1.8$) and better performance, while the static LLM embeddings exhibit low ratios, resulting in rough, difficult-to-model objective landscapes. The failure of static LLM embeddings is therefore a failure of alignment between the representation's geometry and the surrogate model's requirements, motivating a method that can actively learn to resolve this incompatibility.

\subsection{Uncertainty-guided LLM finetuning enables efficient optimization}

To resolve the misalignment between the representation and the surrogate model, the LLM cannot be a static feature extractor; it must learn to adapt. We therefore formalize pretrained LLMs as learnable feature extractors, "taught" by the GP's own probabilistic objective. This approach transforms the LLM from a static component into a dynamic one, tailoring its vast pretrained knowledge to the optimization task at hand.

In the same Buchwald–Hartwig reactions benchmark as above, our joint GP–LLM training substantially improved optimization efficiency (Figure \ref{fig:bh-results}b). An encoder-decoder Google T5 \cite{2020t5} model increased the discovery rate of high-yielding conditions from 24\% to 43\%, with a 79\% relative improvement compared to its static representation. This increase surpassed specialized fingerprints, achieving the new state-of-the-art. 

Similar improvements held across other architecture types. The same approach elevated decoder-only models (61\% relative improvement for Qwen-7B \cite{bai2023qwen}), encoder types (42\% for ModernBERT \cite{modernbert}), and raised the score of closed source OpenAI \cite{openai2024new} embeddings (by 31\%) without the access to the base model's weights. These systematic improvements across major LLM architecture types suggest a general principle rather than an isolated finding: that task-specific adaptation is more important than domain-specific pretraining. Indeed, general-purpose LLMs with no chemistry-specific pretraining (T5 \cite{2020t5}, Qwen-7B \cite{bai2023qwen}) surpassed both chemistry specialized features (DRFP \cite{probst2022reaction}) and specialized chemistry models (T5Chem \cite{christofidellis2023unifying}). 

The mechanism behind the sample-efficiency lies in the training objective. By finetuning the LLM with the GP's marginal likelihood, we merge both components under a shared probabilistic objective that inherently balances data fit with uncertainty estimates. This training paradigm repurposes the uncertainty into an LLM learning signal that guides the reorganization of its latent space. Traditional LLM finetuning has often required thousands of training examples. Here we instead finetune large pretrained language models with 10 initial points (adding one example per iteration in 50 sequential trials) adapting their latent spaces into smooth and calibrated representations for the optimization task.

These results indicate that uncertainty-driven finetuning can unlock task-specific effectiveness from generic LLMs, bypassing the need for specialized model training in each new domain. By unifying natural language with probabilistic calibration, our approach enables the accessibility of language models and the reliability of Bayesian optimization within a single framework.

\subsection{Emergent chemical organization supports interpretable optimization}

\begin{figure}[h!]
\begin{center}
\includegraphics[width=0.99\linewidth]{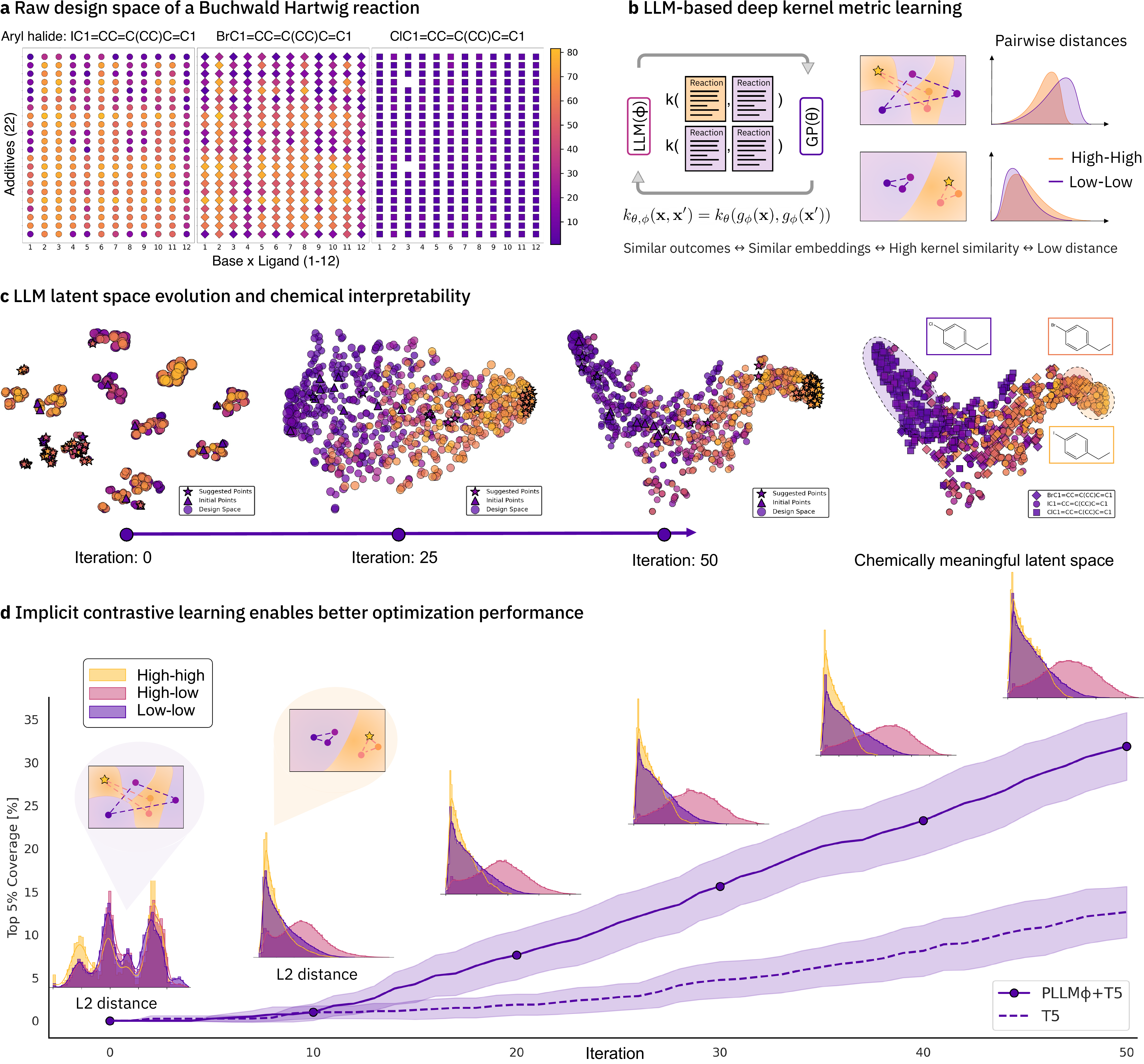}
\end{center}
\caption{\textbf{Implicit contrastive learning and emergent chemical interpretability with LLM-based deep kernel GPs.} \textbf{a}, The raw design space of one of the Buchwald-Hartwig reactions, showing reaction yields across different combinations of additives, bases, and ligands for three different aryl halides (I, Br, Cl). \textbf{b}, A schematic of LLM-based deep kernel metric learning. The joint optimization of the LLM and GP encourages the model to learn embeddings where experiments with similar outcomes position close together in the latent space, resulting in high kernel similarity and low pairwise distance. \textbf{c}, The evolution of the LLM's latent space during the optimization process. Starting from an unstructured state (Iteration 0), the space becomes progressively more organized (Iteration 25), eventually forming a chemically meaningful map where reactions cluster based on key reactivity patterns and performance (Iteration 50). \textbf{d}, The link between implicit contrastive learning and optimization performance. The main plot shows that the adaptive model (PLLM$\phi$+T5) significantly outperforms the static model (T5). The inset histograms show the distribution of pairwise L2 distances between high-yielding and low-yielding experiments at different iterations. As the optimization progresses, the model learns to separate high- and low-performing points, creating a more structured space that enables more efficient discovery.}

\label{fig:contrastive-learning}
\end{figure}

Beyond improved performance, the finetuned LLM’s latent space exhibits emergent organization of chemical knowledge that offers interpretability. A persistent challenge limiting the adoption of AI in experimental science is the 'black box' problem: even if the model's suggestions are effective, they often lack a clear scientific rationale, forcing researchers to follow them blindly rather than with understanding. This prevents the generation of new scientific insights and undermines trust in the optimization process. 

Our framework transforms this paradigm by making the model's reasoning transparent, as uncertainty-guided finetuning fundamentally reorganizes the LLM's latent space into an interpretable map of chemical reality (Figure \ref{fig:contrastive-learning}c). GOLLuM autonomously rediscovers chemical patterns directly from experimental data, organizing the search space into meaningful regions of high and low performance. The reasoning for its suggestions, therefore, lies within the geometry of the latent space itself.

During optimization on the Buchwald-Hartwig tasks, the model automatically learns to cluster experiments according to key chemical features and their outcomes (Figure \ref{fig:contrastive-learning}c). For instance, experiments using high-reactivity iodide-based aryl halides cluster in a region of the latent space associated with high yields, while those using less reactive chlorides occupy a different, lower-performance region. This data-driven grouping mirrors well-known reactivity trends in cross-coupling chemistry, despite the model never being explicitly taught these relationships. This organization emerges progressively, with structured patterns becoming observable by the 10th optimization iteration as the GP marginal likelihood objective acts as an implicit contrastive loss, separating successful and unsuccessful experiments (Figure \ref{fig:contrastive-learning}d).

This emergent structure is functionally critical to the optimizer's efficiency. By grouping successful experiments and distancing failures, the model creates a smoother, more coherent landscape for the GP, enabling more informed and sample-efficient exploration. This clustering transforms the model from a black box into a collaborative tool for scientific discovery. The organized latent space provides a direct rationale for the model’s suggestions, facilitating understanding of why a proposal is likely to succeed based on its proximity to known positive outcomes. The location of a proposed experiment in this latent space carries potentially beneficial information even before running the experiment, enabling new hypotheses and interaction with the optimization process. This interpretability turns a passive cycle of suggestions into an active dialogue between human intuition and machine intelligence, allowing scientists to gain new insights during the optimization rather than solely following outputs.

\subsection{Broad generalizability across diverse chemical domains}

\begin{figure}[h!]
\begin{center}
\includegraphics[width=\linewidth]{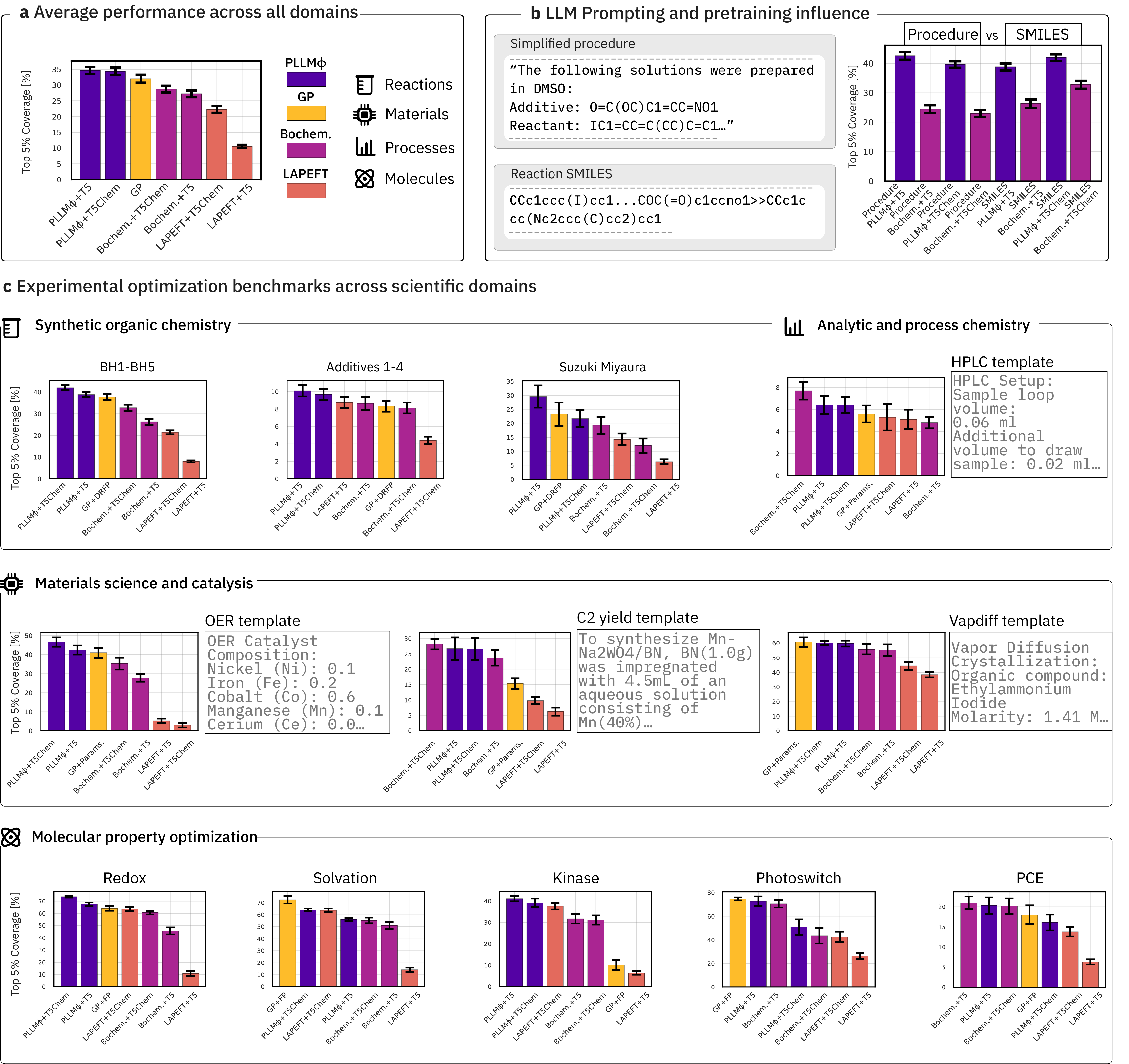}
\end{center}
\caption{\textbf{Benchmarking on various chemistry-related optimization tasks with comparisons to related approaches.} \textbf{a}, On average across all 19 benchmarks, our GOLLuM framework (PLLM$\phi$+T5 or T5Chem) outperforms traditional domain-specific features (GP), static LLM embeddings (Bochemian \cite{rankovic2023bochemian}), and decoupled uncertainty calibration methods (LAPEFT \cite{kristiadi2024sober}). Bars show mean and standard error across 20 seeds per method per benchmark (10 below-median initial points, sequential acquisition-based optimization). \textbf{b}, The framework demonstrates robustness to variations in input format (textual procedure vs. SMILES) and eliminates the need for chemistry-specific pretraining, as the generalist T5 \cite{2020t5} model performs as well as T5Chem \cite{christofidellis2023unifying} \textbf{c}, Detailed results for all 19 optimization tasks, grouped by scientific domains. Synthetic organic chemistry: Buchwald-Hartwig \cite{ahneman2018predicting} (BH1-5), Suzuki Miyaura \cite{perera2018platform} and Ni-catalyzed arylation \cite{prieto2022accelerating} reactions (Additives1-4). Analytic and process chemistry with high-performance liquid chromatography (HPLC setup) \cite{hase_olympus_2021}. Materials science and catalysis:  oxygen evolution reaction catalysts (OER \cite{hase_olympus_2021}), C2 yield optimization \cite{ramos2023bayesian}, vapor diffusion crystallization (Vapdiff) \cite{hase_olympus_2021}. Molecular optimization: minimizing redox potential (Redox) \cite{Agarwal2021} and solvation energy (Solvation) \cite{Agarwal2021}, inhibiting kinase activity (Kinase) \cite{Graff2021}, and maximizing photoswitch absorption wavelengths (Photoswitch) \cite{Griffiths2022a} and power conversion efficiency (PCE) \cite{Lopez2016}. Our method consistently achieves top-tier performance, demonstrating broad applicability from organic synthesis and materials science to molecular property and process optimization. We provide example procedural templates for tasks where SMILES is not applicable.}

\label{fig:benchmarks}
\end{figure}

The central premise of language model-based optimization is the potential for a single framework to leverage prior knowledge across diverse scientific problems without domain-specific feature engineering. Traditional optimization approaches rely heavily on expert-built descriptors tailored to each domain. We hypothesize whether the expressive power of language models under probabilistic alignment could capture the underlying physics and chemistry encoded in textual descriptions, eliminating the need for specialized representations by navigating the optimization from text.

To evaluate this capability, we benchmarked GOLLuM across 19 optimization tasks spanning organic synthesis, materials science, process engineering, and molecular design. These tasks encompassed diverse problem structures, including combinatorial search over discrete molecular spaces, continuous optimization of process conditions, and mixed-variable problems combining both discrete and continuous parameters. Using a single set of fixed hyperparameters without any task-specific modifications, we benchmarked GOLLuM against relevant baselines including specialized feature-based methods, fixed LLM embeddings, and supervised uncertainty approaches (summarized results in Figure \ref{fig:benchmarks}c).

GOLLuM ranked first on average across all 19 benchmarks, consistently identifying optimal or near-optimal solutions with high sample efficiency. The framework maintained robust performance despite cold start initialization and achieved substantial quantitative improvements over competing paradigms ( 8\% over specialized features, 23\% over fixed LLM embeddings, and 114\% over supervised uncertainty methods, Figure \ref{fig:benchmarks}a).

The improvement over specialized features represents a particularly relevant achievement, as these representations typically encode decades of domain expertise and have historically enabled effective optimization in their respective fields. For example, molecular fingerprints capture specific structural motifs known to correlate with chemical properties, while reaction descriptors encode core reactivity patterns. That a language model-based approach can exceed these carefully crafted features suggests that simple textual descriptions of experiments contain richer, more generalizable information than previously recognized.

The 23\% improvement over fixed LLM embeddings validates our hypothesis that optimization-aware finetuning benefits over static representations. Moreover, this approach reduced format dependency that limited fixed embeddings by enabling both the general T5 \cite{2020t5} and T5Chem \cite{christofidellis2023unifying} to achieve comparable performance across both SMILES strings and natural language procedure descriptions (Figure \ref{fig:benchmarks}b). This format robustness facilitates applicability to optimization problems with multiple data representations. In certain cases, however, particularly those involving small numerical values (such as HPLC flow rates of 0.02 mL), finetuning appeared to degrade performance relative to fixed embeddings. This limitation may arise when numerical precision requirements exceed the differentiable information content available in natural language descriptions, highlighting an important boundary condition for text-based optimization approaches.

The highest improvement (114\%) came from integrating uncertainty quantification directly into the training process through Gaussian process-based finetuning, rather than treating uncertainty estimation as a post-hoc learning task. This design ensures that both predictive capacity and uncertainty calibration improve simultaneously, whereas approaches that prioritize the former over the latter may compromise reliability, effectively training models to be good guessers rather than calibrated reasoners \cite{kalai2025language}. (Table \ref{tab:bh_estimates}).

\begin{table}[ht]
\centering
\begin{tabular}{lcc}
\toprule
\textbf{Model} & $R^2$ $\uparrow$ & NLPD $\downarrow$ \\
\midrule
PLLM$\Phi$+T5 (ours)      & $0.6 \pm 0.1$ & $6.9 \pm 1.6$     \\
PLLM$\Phi$+T5Chem (ours) & $0.6 \pm 0.1$ & $8.1 \pm 1.5$     \\
\addlinespace %
GP+DRFP                 & $0.4 \pm 0.3$ & $4.9 \pm 5.0$     \\
\addlinespace
Bochem.+T5              & $0.3 \pm 0.2$ & $38.3 \pm 63.3$   \\
Bochem.+T5Chem          & $0.5 \pm 0.1$ & $49.2 \pm 62.3$   \\
\addlinespace
LAPEFT+T5               & $0.4 \pm 0.1$ & $103.2 \pm 203.4$ \\
LAPEFT+T5Chem           & $0.5 \pm 0.1$ & $357.9 \pm 1016.6$\\
\bottomrule

\end{tabular}
\caption{\textbf{Regression and uncertainty quantification metrics on Buchwald-Hartwig reactions.} We evaluate predictive capabilities by the coefficient of determination ($R^2$, higher is better), and uncertainty calibration by the negative log predictive density (NLPD, lower is better). We report mean and standard deviation values, calculated over 20 independent runs with 60 training points each, using the remaining data for evaluation. The results show that GOLLuM variants (PLLM$\phi$) achieve high predictive accuracy while maintaining strong uncertainty calibration.}
\label{tab:bh_estimates}
\end{table}

\section{Discussion}

In this work, we propose how to resolve the critical trade-off between the domain knowledge of large language models and the principled reliability of Bayesian optimization. By repurposing the uncertainty from a Gaussian process as a direct training signal, we transform a generalist language model into a sample-efficient, domain-aware scientific optimizer. This transformation enables what neither paradigm could achieve alone: the generality of a text-based interface with the reliability required for experimental campaigns.

Our empirical evaluations demonstrate how finetuning language models with Gaussian process uncertainty, consistently matches and outperforms specialized, expert-engineered features in guiding scientific optimization. This robust mechanism allows the framework to generalize effectively across 19 disparate scientific domains without task-specific modifications. Such cross-domain success stems from an emergent understanding of the underlying optimization landscape, a principle quantified by the strong correlation between our geometric smoothness metric ($\ell/\overline{d}$) and search efficiency. This combination of a unified, text-based approach with a reliable probabilistic formulation makes it a powerful tool for accelerating discovery in self-driving laboratories and human-AI collaborative experiments.

While our approach demonstrates broad applicability across the chemical sciences, we acknowledge several limitations that suggest avenues for future research. The cubic scaling of standard GPs may pose a challenge for optimization campaigns with thousands of experiments, suggesting a future need to integrate scalable GP approximations \cite{hensman2015scalable, blei2017variational}. Furthermore, while text provides a powerful universal interface, its effectiveness may be reduced in domains dominated by complex, high-dimensional data lacking a clear linguistic structure, such as protein conformations or crystal structure. Future work could explore hybrid models that combine our framework with specialized encoders for such data types.

This work suggests that a path to more capable and reliable LLMs for science lies in grounding the models in principled probabilistic reasoning. Crucially, our findings reframe uncertainty from a flaw to be mitigated, to a rich signal making language models more reliable and aligned with scientific goals. The ability to use natural language to optimize a chemical reaction, a material synthesis protocol, or a molecular design lowers the barrier to entry for researchers across all scientific disciplines. These results suggest a new standard for AI-guided experimental design: an accessible, interpretable, and reliable toolkit applicable within experimental science. By grounding LLMs in the principled probabilistic objectives, we fundamentally shift their learning goals from making best predictions to providing well-calibrated ones. This approach suggests a different paradigm for achieving specialization, not through more data, but through richer, uncertainty-guided information.

\section{Methods}

\subsection{Bayesian optimization overview}
Bayesian optimization is a sample-efficient method for optimizing black-box functions, potentially expensive to evaluate -- a setting common in chemistry, where each experiment incurs substantial costs. BO works by training a probabilistic surrogate model (typically a GP) on previously observed data and using it to select new, informative queries via an acquisition function. We provide a detailed technical overview in Appendix~\ref{sec:technical-background}. The effectiveness of BO depends on the choice of representations and the quality of the surrogate model -- both of which we improve in this work.

\subsection{Data representation}

BO performance in chemistry is highly sensitive to the choice of data representation, especially given the heterogeneous data types (e.g., categorical reagents, numeric conditions, molecular structures), combinatorial design spaces, and variable numbers of parameters involved -- making representation a critical challenge \citep{rankovic2023bayesian}. Natural language offers a flexible medium for expressing such optimization problems as textual descriptions, while LLMs can transform these inputs -- regardless of type -- into unified continuous embeddings. We construct these embeddings through a two-step process:

\textbf{1. Template construction}: 
   We define each optimization task $t$ as a standardized template:
   $
   t = \text{template}(\{\textit{parameters}, \textit{values}\})
   $
where values define the actual conditions of the problem (e.g., reagents used in a chemical synthesis). For single-variable tasks (e.g., molecular optimization), the template reduces to a single textual identifier such as a molecular SMILES string \citep{anderson1987smiles, weininger1988smiles}. This approach provides a consistent format applicable across a wide range of optimization problems.

\textbf{2. LLM embedding}:
   We process the templated description through LLMs to obtain a fixed-dimensional embedding:
   $
   \mathbf{x} = \text{LLM}(t) \in \mathbb{R}^d
   $. This embedding unifies heterogeneous parameter types and enables compatibility with standard continuous kernels (e.g., Matérn), while preserving inter-parameter relationships and scaling to variable-length inputs.

The resulting embedding vector $\mathbf{x}$ captures both the individual parameter values and their interactions, providing a unified representation for subsequent GP modeling. This approach circumvents the need for designing specialized kernels, as the LLM embedding space naturally encodes meaningful distances -- enabling optimization over mixed categorical and numerical inputs within a continuous space. Moreover, it generalizes to tasks with arbitrary combinations of categorical, numerical or structural parameters -- making it broadly applicable beyond chemistry to domains where design spaces can be expressed through text.

\subsection{Gaussian process with fixed LLM embeddings}

LLM embeddings can be directly used as input vectors to GPs, which model the output based on observed data. In this setup, the embeddings remain fixed throughout the optimization process -- following the approach outlined in BoChemian \cite{rankovic2023bochemian} -- and the model's adaptability comes solely from learning the GP hyperparameters $\theta$. 
We use a GP prior with a Matérn-5/2 kernel with trainable hyperparameters $\theta=\{\ell, \sigma^2, \sigma_n^2, c\}$ representing the lengthscale, signal variance, observation noise variance, and constant mean. This approach relies entirely on the pretrained LLM's embedding space to define input structure -- specifically, the relative positioning of points based on their underlying features.
The GPs with stationary kernels (such as Matérn-5/2) assume this structure reflects meaningful relationships: points close together in the embedding space are expected to have similar outcomes. However, general pretrained LLMs may not reflect chemical similarities and their representations may not encode the right inductive biases for the task. As a result, the GP can struggle to model the objective effectively in the fixed-feature setting, unless the embedding space already captures relevant patterns. This limitation can be addressed through deep kernel methods, which we describe next.

\subsection{Deep kernel Gaussian process}

Deep kernel Gaussian processes \cite{wilson2016deep} combine the flexibility of deep neural networks with the principled uncertainty quantification of Gaussian processes. In this approach, the kernel function is composed with a learned feature transformation:

$$
k_{\text{$\theta, \phi$}}(\mathbf{x}, \mathbf{x}') = k_{\theta}(g_\phi(\mathbf{x}), g_\phi(\mathbf{x}')),
$$

where $g_\phi$ is a parameterized feature extractor with parameters $\phi$. This composition allows the model to learn task-specific feature representations while maintaining the probabilistic properties of the GP framework. The learned transformation and the GP parameters are jointly optimized through the marginal likelihood where $\mathbf{K}_{\theta, \phi}$ is the kernel matrix computed using the transformed features.

\subsection{LLM-based deep kernel}
\label{sec:dkl}
In our framework, we explore different approaches to constructing the feature transformation $g_\phi(\cdot)$.

\textbf{1. Projection layer}: A learned transformation consisting of a linear projection $\mathbf{P} \in \mathbb{R}^{m \times d}$ followed by a non-linear activation function (ELU), applied to fixed LLM embeddings: $g_\phi(\mathbf{x}) = \mathbf{P}\text{LLM}(t)$ where $m$ is the projection dimension. This setup closely follows standard deep kernel learning with a trainable transformation applied on top of fixed features before kernel evaluation. It is particularly useful in settings where LLM weights cannot be accessed, as in the case of closed-source models from OpenAI. The projection layer learns to emphasize or suppress different aspects of fixed LLM embeddings, effectively creating a task-specific representation.

\textbf{2. PEFT-adapted LLM}: Low-rank adaptation of LLM parameters: $g_\phi(\mathbf{x}) = \text{LLM}\phi(t)$
where $\phi$ represents the trainable adapter parameters. Parameter efficient finetuning (PEFT) \citep{houlsby2019parameter, hu2022lora, li2023loftq} addresses the challenge of adapting large language models by updating a smaller (often several orders of magnitude fewer) number of parameters, typically inserted into or alongside the LLM architecture. We employ Low-Rank Adaptation (LoRA) \citep{hu2022lora} to preserve potential chemical knowledge captured during pretraining and learn task-specific adaptations, while avoiding catastrophic forgetting or compromising general capabilities.

\textbf{3. Combined approach}: Sequential application of LoRA and projection: $g_\phi(\mathbf{x}) = \mathbf{P}\text{LLM}\phi(t)$, thus combining the benefits of both worlds. The LoRA adapters allow the LLM to adapt its internal representations to the optimization task, while the projection layer provides an additional degree of freedom to reshape the embedding space.

With any of these methods, we optimize the parameters $\phi$ (projection matrix and/or LoRA parameters) jointly with the GP hyperparameters through the marginal likelihood. In other words, we are finetuning the LLM through the GP loss which allows the model to learn transformations that both preserve relevant chemical information, organize the latent space to better reflect the structure of the optimization objective, and provide well-calibrated uncertainty measures. 

\subsection{LLM finetuning as GP marginal likelihood optimization}

Let $\mathcal{L}(\theta, \phi)$ denote the GP marginal likelihood of observing targets $\mathbf{y}$ given inputs $\mathbf{X}$, LLM parameters $\phi$, and GP hyperparameters $\theta$:

\begin{equation}
\mathcal{L}(\theta, \phi) = \log p(\mathbf{y}|\mathbf{X}, \theta, \phi) = -\frac{1}{2}(\mathbf{y}^\top\mathbf{K}_{\theta,\phi}^{-1}\mathbf{y} + \log|\mathbf{K}_{\theta, \phi}| + n\log2\pi).
\label{eq:likelihood}
\end{equation}

To optimize the embedding parameters $\phi$ jointly with the GP hyperparameters $\theta$, we maximize the marginal likelihood using gradient-based optimization:
\begin{equation}
    \theta^*, \phi^* = \arg\max_{\theta,\phi}\mathcal{L}(\theta,\phi).
\end{equation}
We compute the gradients of the marginal likelihood with respect to the parameters via standard backpropagation:
\begin{equation}
    \nabla_{\theta,\phi}\mathcal{L}(\theta,\phi) 
    = \frac{1}{2}\mathbf{y}^\top \mathbf{K}_{\theta,\phi}^{-1}\left(\nabla_{\theta,\phi}\mathbf{K}_{\theta,\phi}\right)\mathbf{K}_{\theta,\phi}^{-1}\mathbf{y} - \frac{1}{2}\text{Tr}\left(\mathbf{K}_{\theta,\phi}^{-1}\nabla_{\theta,\phi}\mathbf{K}_{\theta,\phi}\right).
\label{eq:gradients}
\end{equation}
We perform the joint optimization with separate learning rates for embedding parameters ($\phi$) and GP hyperparameters ($\theta$) to encourage stable convergence and avoid overfitting of either component.

\subsection{Implicit metric learning}
The explicit feature of GPs to evaluate the similarities in the output based on the distances in the input space creates a contrastive learning effect for LLM embeddings. This beneficial consequence arises from the two-fold utilization of the GP marginal likelihood. The kernel function $ k_{\text{$\theta, \phi$}}(\mathbf{x}, \mathbf{x}') = k_{\theta}(g_\phi(\mathbf{x}), g_\phi(\mathbf{x}'))$ measures similarity between points, and optimizing the marginal likelihood (Eq. \ref{eq:likelihood}) encourages embedding distances to decrease between points with similar outputs and increase between points with dissimilar outputs. The contrastive learning effect comes directly from the GP marginal likelihood optimization. For a kernel based on distances (like Matérn) we can rewrite the term $\mathbf{y}^\top\mathbf{K}_{\theta,\phi}^{-1}\mathbf{y}$ as a weighted sum of pairwise interactions (with weights $w_{ij}$ defined by the inverse kernel matrix) inducing implicit contrastive learning objective $\mathcal{L}_\text{implicit}$:

\begin{equation}
\mathcal{L}_\text{implicit}(\theta, \phi) \propto \sum_{i,j} w_{ij} \cdot \|g_\phi(\mathbf{x}_i) - g_\phi(\mathbf{x}_j)\|_2, \quad 
\begin{cases}
\|g_\phi(\mathbf{x}_i) - g_\phi(\mathbf{x}_j)\|_2 \downarrow \text{ if } \|y_i - y_j\| \text{ is small} \\
\|g_\phi(\mathbf{x}_i) - g_\phi(\mathbf{x}_j)\|_2 \uparrow \text{ if } \|y_i - y_j\| \text{ is large}
\end{cases}
\label{eq:implicit}
\end{equation}

In other words, the joint GP optimization induces high kernel values (small distances) between points with similar outputs and low kernel values (large distances) between points with different outputs, therefore separating the embedding space into distinct categories. This reorganization in the latent space happens automatically through the optimization of the deep kernel parameters, adapting the feature space to better align with outcomes without requiring explicit contrastive loss terms. 

\subsection{Joint optimization and computational cost}

For deep kernel models, we jointly optimize the GP and LLM parameters by maximizing the marginal likelihood. We use the AdamW optimizer with separate learning rates for the GP hyperparameters (0.2) and the trainable LLM weights (0.002). For larger models, we run experiments on an NVIDIA H100 GPU. LoRA keeps the method lightweight; for T5, fewer than 0.2\% of the total model weights are updated during training. We provide full implementation details, including PEFT configurations, learning rate schedules, and hardware specifications, in Supplementary Information section \ref{sup:implementation}.

\subsection*{Acknowledgments}
We would like to acknowledge Joshua Sin and Edvin Fako for their invaluable insights, feedback, and constructive discussions throughout this study.

This publication was created as part of NCCR Catalysis (grant number 225147), a National Centre of Competence in Research funded by the Swiss National Science Foundation.

\subsection*{Code and data availability}
Code and data available at \href{https://github.com/schwallergroup/gollum}{https://github.com/schwallergroup/gollum}.

\bibliographystyle{naturemag}  
\bibliography{iclr2025_conference}

\appendix

\section{Related work}
Adapting LLMs to specialized tasks through finetuning typically optimizes for predictive accuracy \cite{jablonka2023gpt,xie2023darwin}. Such domain adaptation neglects the dimension of epistemic uncertainty to indicate when model outputs should not be trusted. In this context, the challenge of extracting reliable uncertainties from LLMs for efficient BO has introduced diverse approaches that fall into four categories.

(1) Prompt-based methods like BOLIFT \citep{ramos2023bayesian}, estimate uncertainty by aggregating multiple LLM responses, while LLAMBO and LLM-GO \citep{nguyen2024predicting, macknight2025pre} query LLMs as optimizers. (2) Embedding-based methods, such as BoChemian \citep{rankovic2023bochemian}, model GPs on static LLM representations. (3) Surrogate conversion models either predict uncertainty through pretrained in-context regressor \cite{nguyen2024predicting} or transform LLMs into Bayesian neural networks \cite{kristiadi2024sober} through PEFT and Laplace approximation (LAPEFT \citep{laplace2021} following Bayesian LoRA approach \cite{yang2023bayesian}). (4) Hybrid approaches like BOPRO \citep{agarwalsearching} combine fixed LLM embeddings for GP-based acquisition optimization with prompt-conditioned LLM generation, using in-context examples to explore the solution space. BORA \cite{cisse2025language} uses a heuristic policy to regulate LLM's high-level reasoning and its suggestions for new regions of the search space to explore. While promising, these methods face limitations from heavy prompt engineering to post-hoc uncertainty fixes. Our approach offers a solution by integrating uncertainty modeling \textit{during} training within both the LLM and the surrogate model (GP) through LLM-based deep kernel strategy.

Previous works on integrating LLMs and GPs \citep{rankovic2023bayesian, ramos2023bayesian, rankovic2023bochemian, kristiadi2024sober} for chemical optimization primarily use LLMs as fixed feature extractors. Despite LLM's flexibility in converting diverse parameters into fixed-dimensional representations, these approaches essentially reduce their power to sophisticated encoding tools. The static embeddings struggle to match, let alone surpass, carefully engineered domain-specific features, creating an artificial barrier between LLM expressiveness and GP rigor. 

The closest related approach to ours is the one by Kristiadi et al. \citep{kristiadi2024sober}. It incorporates supervised LLM finetuning during optimization (for better prediction), followed by post-hoc Laplace approximation of learned weights (for uncertainty estimates). The sequential approach however, decouples uncertainty quantification from the learning process and optimizes for prediction accuracy (lower MSE loss) rather than optimization itself. Our method fundamentally differs by integrating LLMs directly into the GP framework as a deep kernel, making uncertainty quantification an integral part of the optimization objective through the GP marginal likelihood.

\begin{figure}[h!]
\begin{center}
\includegraphics[width=\linewidth]{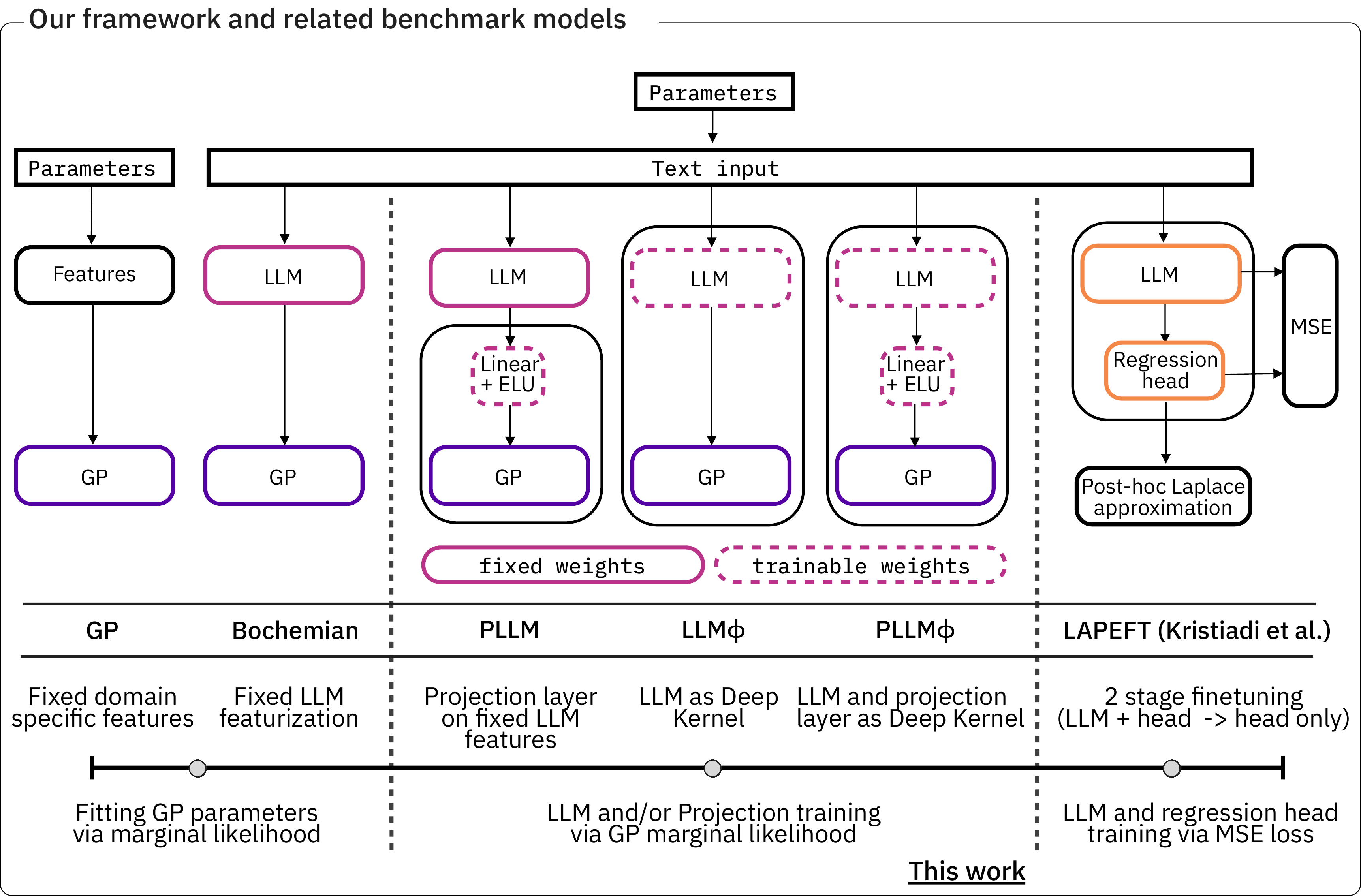}
\end{center}
\caption{\textbf{Architectural comparison of our proposed framework and related benchmark models.} The diagram illustrates the data and training flow for each approach. From left to right: A standard GP baseline using fixed domain-specific features. Bochemian \cite{rankovic2023bochemian}, which uses a fixed LLM as a feature extractor. Our three proposed variants (PLLM, LLM$\phi$, and PLLM$\phi$), which are distinguished by their trainable components (a projection layer, the LLM itself, or both). Critically, our framework uses the GP's marginal likelihood to jointly train these components. This is contrasted with LAPEFT \cite{kristiadi2024sober}, which trains an LLM and regression head via an MSE loss and estimates uncertainty as a separate, post-hoc step.}
\label{fig:architectures}
\end{figure}

Existing deep kernel methods \citep{singh2024deep, chen2022meta} in chemistry constrain to using graph neural network kernels and domain specific representations. Advancing this concept, we demonstrate that LLM-based kernels provide richer embedding space and easier adaptation to various domains. In the latent space optimization, our approach relates to BO with variational autoencoders (VAEs). For example, \citep{griffiths2020constrained} use VAE-BO for constrained optimization of molecules while \citep{grosnit2021high} introduce explicit deep metric learning to structure the latent space. We elevate this approach by replacing pretrained VAEs with general purpose LLMs. Moreover, our joint optimization induces implicit metric learning, directly structuring the latent space without the need for explicit contrastive objectives. Importantly, deep kernel learning in low-data regime has long been considered challenging due to issues like overfitting and training instability \cite{ober2021promises}, with some approaches opting for coordinate ascent or bi-level optimization strategies to mitigate these issues \citep{stanton2022accelerating, chen2022meta}. In contrast, we build a robust and stable training pipeline using standard backpropagation with separate learning rates for the LLM and GP components, avoiding the need for complex optimization schedules or custom regularization.

Operating in high-dimensional BO (HDBO) spaces (768 features with BERT-based models, 4096 with larger decoder types), our work demonstrates successful optimization that challenges conventional assumptions about HDBO limitations \citep{frazier2018tutorial}. Through our empirical analysis, we identify key principles for effective HDBO: embeddings should structure following the GP's inductive bias. This organization of the embedding space plays a crucial role in overcoming dimensionality challenges that have traditionally constrained BO approaches. Recent work on HDBO has identified vanishing gradients in GP training and acquisition optimization as a key challenge, and proposed solutions based on informed lengthscale priors and local search heuristics \citep{papenmeier2025understanding}. Our approach complements this line of work by focusing on the structure of the representation space itself, showing that smoothness and generalization can emerge naturally when the embedding is well-aligned with the GP kernel. Our observations contribute to emerging approaches in understanding and improving HDBO performance \citep{papenmeier2025understanding, xu2025standard}.

Within chemistry, high-dimensional features are the default (one-hot encodings \citep{chuang2018comment}, fingerprints \citep{rogers2010extended, schneider2015development, capecchi2020one, probst2022reaction, schwaller2021mapping} quantum mechanical descriptors \citep{ahneman2018predicting, shields2021bayesian}) but their selection remains challenging \citep{rankovic2023bayesian}. We overcome this problem through unified representation learning as an implicit objective of the optimization strategy itself. As a result, we achieve adaptive features for any optimization at hand, without requiring traditionally used expert-based descriptors.

\section{Technical background}
\label{sec:technical-background}
\subsection{Bayesian optimization}

Bayesian optimization (BO) is a suitable method for optimizing expensive-to-evaluate functions with unknown analytic form or gradients. Such problems are common in chemistry where experimental evaluations are often costly, time-consuming, and only available through real-world lab experiments. The primary objective of BO is to find:

\begin{equation}
\mathbf{x}^* = \arg\max_{\mathbf{x} \in \mathcal{X}} f(\mathbf{x})
\end{equation}

where $f: \mathcal{X} \rightarrow \mathbb{R}$ is the objective function over domain $\mathcal{X}$. In practical chemistry applications, we often work in a constrained domain $\mathcal{X}_{\text{pool}}$ with limited set of possible experimental conditions or molecular structures (e.g., feasible reaction conditions, available reagents or compound libraries). The optimization objective may involve maximizing reaction yield or selectivity, or minimizing properties such as reaction time, cost or toxicity.

\subsection{Sequential decision process}
Bayesian optimization operates as a sequential decision-making process that balances exploration and exploitation. Key components include (1) a probabilistic surrogate model of the underlying objective function $f$ and (2) an acquisition function $\alpha$. The acquisition function guides the optimization process by proposing subsequent evaluation points. Common choices include expected improvement (EI \cite{jones1998efficient}), probability of improvement (PI \cite {jones2001taxonomy}), upper confidence bound (UCB \cite{auer2002using}, \cite{srinivas2009gaussian}), and Thompson sampling (TS \cite{thompson1933likelihood}). Acquisition function selection over the points in the design space relies on the surrogate model and its predictive and uncertainty estimates. For example EI selects points that, in expectation, improve upon the current best observed value $f(\mathbf{x}_{\text{best}})$:
\begin{equation}
\alpha_{\text{EI}}(\mathbf{x}|\mathcal{D}_t) = \mathbb{E}_{p(f|\mathcal{D}_t)}[\max(f(\mathbf{x}) - f(\mathbf{x}_{\text{best}}), 0)]
\end{equation} 
In that sense, the choice of a surrogate model is critical to the success of BO. Gaussian Processes (GPs) are the most common choice due to their flexibility and ability to quantify uncertainty, making them particularly suitable for guiding the exploration of vast chemical spaces while operating in low-data regimes.

\subsection{Gaussian processes and marginal likelihood optimization}

GPs provide a flexible non-parametric method for modeling unknown functions. A GP places a prior distribution
\begin{equation}
f(\mathbf{x}) \sim \mathcal{GP}(c, k(\mathbf{x}, \mathbf{x}')),
\end{equation}

defined by a mean function \( c \) (typically 0 or constant) and a kernel function \( k \) encoding pairwise similarity between inputs and prior assumptions about function smoothness and variability. A common choice is the Matérn-5/2 kernel

\begin{equation}
k_{\text{Matérn-5/2}}(\mathbf{x}, \mathbf{x}') = \sigma^2 \left(1 + \frac{\sqrt{5} d}{\ell} + \frac{5d^2}{3\ell^2} \right) \exp\left(-\frac{\sqrt{5} d}{\ell} \right),
\end{equation}

where \( d = \| \mathbf{x} - \mathbf{x}' \|_2 \), \(\ell\) is the lengthscale, and \(\sigma^2\) is the signal variance.

Given training data \(\mathcal{D} = \{(\mathbf{x}_i, y_i)\}_{i=1}^n\), the GP posterior allows closed-form prediction of the function at new points, along with uncertainty estimates. Crucially, the GP hyperparameters $\theta=\{c, \ell, \sigma^2, \sigma_n^2\}$ are learned by maximizing the marginal likelihood of the data:

\begin{equation}
\mathcal{L}(\theta) = \log p(\mathbf{y}|\mathbf{X}, \theta) = -\frac{1}{2}(\mathbf{y}^\top\mathbf{K}_{\theta}^{-1}\mathbf{y} + \log|\mathbf{K}_{\theta}| + n\log2\pi)
\label{eq:gp-likelihood}
\end{equation}

where \(\mathbf{K}_\theta\) is the kernel matrix incorporating observation noise \(\sigma_n^2\), evaluated on the training inputs using kernel parameters ($\ell, \sigma^2$). If a constant mean function $c$ is used, the targets $\mathbf{y}$ are centered as 
$\mathbf{y}-c\mathbf{1}$ during marginal likelihood computation. In the standard fixed-feature setting, input \(\mathbf{x}\) is mapped to a feature vector via a static transformation (e.g., molecular fingerprints or frozen LLM embeddings), and the GP operates solely on these representations. The optimization updates \(\theta\), adapting the GP’s inductive bias to the fixed feature space.

\subsection{Deep kernel Gaussian processes}

Deep kernel Gaussian processes (DKGPs) introduce an additional parameter set \(\phi\) to the optimization objective by integrating neural network-based feature transformations into the GP kernel. Formally, the kernel function becomes:
\[
k_{\theta,\phi}(\mathbf{x}, \mathbf{x}') = k_\theta(g_\phi(\mathbf{x}), g_\phi(\mathbf{x}')),
\]
where \(g_\phi(\cdot)\) is a learned data representation parameterized by \(\phi\). This formulation enables the model to adapt the input space to the task at hand while preserving the uncertainty modeling properties of the GP.

The transformation \(g_\phi\) can take the form of any neural architecture suitable for the data modality. The original paper applied the DKGP architecture on regression tasks with images using convolutional neural networks \cite{wilson2016deep}, while subsequent works have extended it to structured chemical domains using graph neural networks \cite{singh2024deep}. As detailed in the main section, we apply this framework to textual chemical representations by using large language models (LLMs) as the deep kernel feature extractor. This allows us to incorporate both pretrained domain knowledge and task-specific adaptation within the BO loop.

\subsection{Large language models}

LLMs process textual inputs by converting them into dense vector representations through a sequence of tokenization \cite{}, embedding \cite{} and attention-based \cite{} transformations. Tokenization involves the process of splitting the input text into subword units (tokens) using a model-specific vocabulary (e.g., SentencePiece \cite{kudo2018sentencepiece}, Byte-Pair Encoding \cite{radford2019language}). The tokens are mapped to continuous vectors via learned embedding layers and passed through multiple self-attention layers that capture contextual relationships between tokens.

LLMs can follow different architectural designs: encoder-only (e.g., BERT \cite{devlin2018bert}), decoder-only (e.g., Qwen \cite{bai2023qwen}), and encoder-decoder (e.g., T5 \cite{2020t5}). Encoder-based models process the full input bidirectionally and are suited for classification and regression \cite{}. Decoder-only models generate text autoregressively with causal masking \cite{}. Encoder-decoder models combine both components and are often used for sequence-to-sequence tasks \cite{}. The architecture choices impact the structure and pooling strategies used to extract unified representations from the variable-length token sequences.

Pooling refers to the process of aggregating a sequence of token-level representations produced by a language model into a single fixed-dimensional embedding. Encoder-based models often use the hidden state corresponding to the special [CLS] token or apply mean-pooling across token embeddings. Decoder-only models typically use the final hidden state of the last non-padding token. For encoder-decoder models, pooling is applied over the encoder-side hidden states.

\subsection{Parameter-efficient LLM finetuning}
Although pretrained LLM embeddings encode rich semantic information, they are not tailored to specific downstream tasks. In that sense, adapting LLMs through finetuning allows for better task-specific capabilities. However, updating LLM weights can be computationally prohibitive due to their large size (often billions of parameters in modern LLMs). Parameter-efficient finetuning (PEFT), however, provides a recipe for LLM task alignment by adapting a smaller subset of parameters while leaving the majority of the model unchanged. 

One such approach is Low-Rank Adaptation (LoRA) \cite{hu2022lora}, which injects trainable low-rank matrices into existing weight layers. Instead of updating a weight matrix \( W \in \mathbb{R}^{d \times k} \), LoRA learns a low-rank update of the form:
\[
\Delta W = A B, \quad \text{where } A \in \mathbb{R}^{d \times r}, \quad B \in \mathbb{R}^{r \times k}, \quad r \ll \min(d, k)
\]
The adapted weight becomes \( W' = W + \Delta W \), allowing task-specific learning with a parameter count that scales with \(r\), the rank of the decomposition. This method allows efficient finetuning and mitigates the risk of catastrophic forgetting by preserving the pretrained weights.

\subsection{Pseudocodes}

\begin{algorithm}[h]
\caption{Constrained Bayesian optimization}
\label{alg:bo}
\begin{algorithmic}[1]
\REQUIRE Initial dataset $\mathcal{D}_0 = \{(\mathbf{x}_i, y_i)\}_{i=1}^{n_0}$, candidate pool $\mathcal{X}_{\text{pool}}$, budget $T$, objective function $f$
\STATE Initialize surrogate model (e.g., GP) using $\mathcal{D}_0$
\FOR{$t = 1$ to $T$}
    \STATE Fit surrogate model to current data $\mathcal{D}_{t-1}$
    \FORALL{$\mathbf{x} \in \mathcal{X}_{\text{pool}}$}
        \STATE Compute acquisition value $\alpha(\mathbf{x} \mid \mathcal{D}_{t-1})$
    \ENDFOR
    \STATE Select next point: $\mathbf{x}_t = \arg\max_{\mathbf{x} \in \mathcal{X}_{\text{pool}}} \alpha(\mathbf{x} \mid \mathcal{D}_{t-1})$
    \STATE Evaluate objective function: $y_t = f(\mathbf{x}_t)$
    \STATE Update dataset: $\mathcal{D}_t = \mathcal{D}_{t-1} \cup \{(\mathbf{x}_t, y_t)\}$
    \STATE Remove $\mathbf{x}_t$ from $\mathcal{X}_{\text{pool}}$
\ENDFOR
\RETURN Best observed point: $\mathbf{x}^* = \arg\max_{(\mathbf{x}, y) \in \mathcal{D}_T} y$
\end{algorithmic}
\end{algorithm}

\begin{algorithm}[h]
\caption{Bayesian optimization with LLM-based deep kernel GP}
\label{alg:bo-llmgp}
\begin{algorithmic}[1]
\REQUIRE Initial dataset \(\mathcal{D}_0 = \{(\mathbf{x}_i, y_i)\}_{i=1}^{n_0}\), candidate pool \(\mathcal{X}_{\text{pool}}\), budget \(T\)
\FOR{$t = 1$ to $T$}
    \STATE Initialize parameters \(\phi\) (LLM) and \(\theta\) (GP)
    \STATE \textbf{Train LLM-GP model:}
    \REPEAT
        \STATE Compute embeddings: \(\mathbf{z}_i = g_\phi(\mathbf{x}_i)\) for all \((\mathbf{x}_i, y_i) \in \mathcal{D}_{t-1}\)
        \STATE Evaluate GP marginal log-likelihood \( \log p(\mathbf{y} \mid \mathbf{z}, \theta) \)
        \STATE Update \(\phi, \theta\)
    \UNTIL{convergence}
    \STATE \textbf{Compute acquisition on candidate pool:}
    \FORALL{\(\mathbf{x}_j \in \mathcal{X}_{\text{pool}}\)}
        \STATE \(\mathbf{z}_j = g_\phi(\mathbf{x}_j)\)
        \STATE Compute \(\alpha(\mathbf{z}_j; \theta)\)
    \ENDFOR
    \STATE Select next input: \(\mathbf{x}_t = \arg\max_{\mathbf{x}_j} \alpha(\mathbf{z}_j)\)
    \STATE Observe outcome: \(y_t = f(\mathbf{x}_t)\)
    \STATE Update dataset: \(\mathcal{D}_t = \mathcal{D}_{t-1} \cup \{(\mathbf{x}_t, y_t)\}\)
    \STATE Remove \(\mathbf{x}_t\) from \(\mathcal{X}_{\text{pool}}\)
\ENDFOR
\RETURN Best input: \(\arg\max_{(\mathbf{x}, y) \in \mathcal{D}_T} y\)
\end{algorithmic}
\end{algorithm}

\section{LLM and domain-specific representations}

\subsection{Molecular Representations}
Molecular representations have been extensively studied in chemistry, leading to (1) fingerprints \citep{rogers2010extended, cereto2015molecular, capecchi2020one}, (2) Simplified Molecular Input Line Entry System (SMILES) strings \citep{anderson1987smiles, weininger1988smiles}, (3) molecular graph-based features \citep{kearnes2016molecular} or (4) more physics-informed descriptors derived from electronic structure calculations \cite{bannwarth2019gfn2}. Each of these representations encodes different aspects of molecular structure and properties making their utility task-dependent.

For example, molecular fingerprints can be effective for tasks involving structural similarity or substructure-driven properties, while quantum-derived features may be better suited for tasks involving electronic properties. For applications in BO, these representations typically require specialized kernel functions to capture relevant similarity \citep{griffiths2023gauche}.

To compare domain-specific representation to general LLM-based ones, we set molecular fingerprints as input to a GP in all molecular property optimization benchmarks.

\subsection{Reaction representations}
Chemical reactions, on the other hand, attach an additional layer of complexity beyond molecular representation including reaction conditions and procedural descriptions. In that sense, they present a unique challenge for machine learning due to an increased complexity and heterogeneous nature. Reaction conditions typically comprise multiple parameter types: numerical values (temperature, concentration, time), categorical variables (catalyst type, solvent choice), and detailed procedural descriptions, making their featurization challenging. Reaction representations used in ML range from simple one-hot encodings \citep{chuang2018comment} to more elaborate reaction fingerprints \citep{rogers2010extended, schneider2015development, capecchi2020one, probst2022reaction}, quantum mechanical descriptors \citep{ahneman2018predicting, shields2021bayesian} and learned representations \citep{schwaller2021mapping}. 

We make extensive use of Differential Reaction Fingerprints (DRFPs) \citep{probst2022reaction}, previously shown to achieve state-of-the-art results on reaction optimization tasks compared to a variety of molecular and reaction descriptors \cite{rankovic2023bayesian}. Generated by first computing circular fingerprints for each reactant and product and then taking their symmetric difference, DRFPs highlight the structural changes during the reaction while remaining computationally cheap. We input DRFPs to GP in reaction optimization tasks with available reaction SMILES. For other chemical optimization tasks we generate features through one-hot encoding of categorical variables, concatenated with numerical parameter values.

\subsection{LLM representations}
\subsubsection{\textbf{Template construction}}
We define each task $t$ through a standardized template:
$
t = \text{template}(\{\text{parameters}, \text{values}\})
$
where the template converts various parameter types into a structured text format:
\begin{verbatim}
The reaction was prepared with:
temperature: {numerical_value}°C
solvent: {solvent_smile}
ligand: {ligand_smile}
\end{verbatim}

\subsubsection{Encoder-based models}

\textbf{Encoder-based}

Encoder-based language models, originally developed for natural language understanding tasks, have long been used to generate fixed-dimensional vector representations of text. These models, typically based on the transformer encoder architecture, process input sequences bidirectionally. In that sense, they have been widely adopted in various downstream tasks such as classification, clustering, and semantic similarity.

With the recent evolution toward larger-scale pretraining, encoder-only models have also followed the trajectory of large language models, yielding high-capacity embedding models suitable for diverse domains beyond natural language. These modern embedding models are trained on massive corpora with contrastive or retrieval-oriented objectives, making them particularly effective for extracting general-purpose sentence and document embeddings.

In this work, we evaluate three encoder-based embedding models on tasks of representing chemical procedures and reaction descriptions for BO.

\textbf{ModernBERT \cite{modernbert}}: A compact and efficient embedding model trained with a retrieval objective, designed for high-speed and high-quality sentence representations.

\textbf{UAE-Large  \cite{li2023angle}}: The Universal Alignment Embedding model, trained for multilingual and multimodal generalization with a strong emphasis on alignment across domains.

\textbf{MXBAI-Embed  \cite{emb2024mxbai}}: A large-scale embedding model from Mixedbread-AI, trained to preserve semantic similarity across a broad range of tasks, including code, math, and text.

\subsubsection{Encoder-decoder models}

Encoder-decoder architectures, such as the T5 family \citep{2020t5}, consist of two transformer modules: an encoder that processes the input sequence and a decoder that generates output sequences, typically in an autoregressive fashion. For embedding tasks, representations are typically extracted from the encoder side, which embeds the input text into a fixed-length latent representation. Compared to encoder-only models, encoder-decoder architectures are often pretrained with sequence-to-sequence objectives such as masked span prediction or denoising, making them well-suited for tasks involving paraphrasing, summarization, or input–output alignment.

We evaluate the encoder outputs from three encoder-decoder models:

\textbf{T5 (base variant) \citep{2020t5}}: A widely-used general-purpose model pretrained on a multi-task mixture of unsupervised and supervised NLP tasks. We use the encoder outputs as text embeddings.
    
\textbf{T5Chem \citep{christofidellis2023unifying}} : A domain-adapted variant of T5, finetuned on chemical tasks using the GT4SD framework. It is trained on a multitask mixture involving molecular property prediction, retrosynthesis, and chemical text modeling, making it more specialized for chemistry-related input sequences.
    
\textbf{Instructor \citep{su2022one}}: An instruction-tuned encoder-decoder model trained on natural language–task pairs. It learns to produce embeddings guided by a task description (e.g., "Represent the reaction for similarity search"), making it suitable for alignment-sensitive downstream applications.

\subsubsection{Decoder-only models}

Decoder-only architectures, exemplified by models in the GPT \cite{radford2018improving} family, generate outputs autoregressively by predicting each token conditioned on all previous ones. While traditionally used for generation tasks, these models can also produce dense representations of input text by extracting hidden states from specific tokens (e.g., the final token or special marker tokens). Decoder-only models are typically pretrained with causal language modeling objectives and operate unidirectionally, which distinguishes their contextual encoding behavior from encoder-based models.

In this work, we evaluate several decoder-style models for embedding chemical procedures and reaction descriptions:

\textbf{OpenAI embeddings \citep{openai2024new}} A widely-used commercial API that provides text embeddings via proprietary transformer models. While the architectural details are not public, we assign them to decoder-style GPT family.

\textbf{Qwen2-7B-Instruct \citep{bai2023qwen}} A large-scale instruction-tuned language model from Alibaba, based on a decoder-only architecture. We use this model in embedding mode by extracting the hidden state of the last non-padding token.

\textbf{GTE-Qwen2-7B-Instruct \citep{li2023towards}} A retrieval-optimized variant of Qwen2, finetuned to produce sentence-level embeddings with improved performance on similarity and ranking tasks.

\textbf{LLaMA 3–8B \citep{grattafiori2024llama}}  Meta’s open LLaMA 3 model in its original instruction-tuned form, without additional adaptation for embeddings.

\textbf{LLM2Vec models \citep{behnamghader2024llm2vec}} We also evaluate decoder-only LLMs adapted for embedding tasks using the LLM2Vec framework. These models, such as LLM2Vec–Meta-Llama-3 and LLM2Vec–Mistral-7B, are trained with masked next token prediction (MNTP) to enable bidirectional context modeling and use supervised mean pooling over selected internal layers. This adaptation allows decoder-only transformers to behave similarly to encoder models in embedding quality, while preserving their original architecture.

\subsection{Representations overview}
We build LLM representations with models from HuggingFace (HF) selected through their base architecture and the results on MTEB in summarization task. In Table \ref{tab:repr-comparison} we give an overview of all representations used in this paper alongside specifics on the dimensionality, pooling, architecture, pretraining, and connections to chemistry. We also include HF sources for LLM-based representations and links to chemistry-related molecular fingerprints and DRFP featurization methods.

\begin{table*}[h]
\centering
\small
\begin{tabularx}{\textwidth}{l l l X l c l}
\toprule
\textbf{Model} & \textbf{Arch.} & \textbf{Pretrain. obj.} & \textbf{Pool} & \textbf{Chem.} & \textbf{Dim.} & \textbf{Source} \\
\midrule
\multicolumn{7}{l}{\textit{Molecular}} \\
Fingerprints & / &  / & / & Yes & 2048 & \href{https://www.rdkit.org/docs/source/rdkit.Chem.rdMolDescriptors.html\#rdkit.Chem.rdMolDescriptors.GetMorganFingerprint}{Morgan} \\
\multicolumn{7}{l}{\textit{Reaction}} \\
DRFP & / &  / & / & Yes & 2048 & \href{https://github.com/reymond-group/drfp}{DRFP} \\
\multicolumn{7}{l}{\textit{Encoder}} \\
ModernBERT & Encoder & Retr. contrastive & CLS & No & 768 & \href{https://huggingface.co/nomic-ai/modernbert-embed-base}{HF} \\
MXBAI-Embed & Encoder & General-purpose & CLS & No & 1024 & \href{https://huggingface.co/mixedbread-ai/mxbai-embed-large-v1}{HF} \\
UAE-Large & Encoder & Align./Multimod. & CLS & No & 1024 & \href{https://huggingface.co/WhereIsAI/UAE-Large-V1}{HF} \\
\midrule
\multicolumn{7}{l}{\textit{Enc-Dec}} \\
T5-Base & Enc-Dec & Mask span pred. & Mean & No & 768 & \href{https://huggingface.co/google-t5/t5-base}{HF} \\
T5Chem & Enc-Dec & Chem multitask & Mean & Yes & 768 & \href{https://huggingface.co/GT4SD/multitask-text-and-chemistry-t5-base-standard}{HF} \\
Instructor & Enc-Dec & Instr. align. & Weigh. Mean & Part.\textsuperscript{$\dagger$} & 768 & \href{https://huggingface.co/hkunlp/instructor-xl}{HF} \\
\midrule
\multicolumn{7}{l}{\textit{Decoder}} \\
OpenAI Embedding & Decoder\textsuperscript{$\ast$} & Proprietary & Unk. & Unk. & 3072 & \href{https://openai.com/index/new-embedding-models-and-api-updates/}{OpenAI} \\
Qwen2-7B-Instruct & Decoder & Instr. tuning & Last & No & 3584 & \href{https://huggingface.co/Qwen/Qwen2-7B-Instruct}{HF} \\
GTE-Qwen2 & Decoder & Contrastive retr. & Last & No & 3584 & \href{https://huggingface.co/Alibaba-NLP/gte-Qwen2-7B-instruct}{HF} \\
LLM2Vec–LLaMA3 & Decoder & Supervised pool. & Last & No & 4096 & \href{https://huggingface.co/McGill-NLP/LLM2Vec-Meta-Llama-3-8B-Instruct-mntp-supervised}{HF} \\
LLM2Vec–Mistral & Decoder & Supervised pool. & Last & No & 4096 & \href{https://huggingface.co/McGill-NLP/LLM2Vec-Mistral-7B-Instruct-v2-mntp-supervised}{HF} \\
LLaMA 3–8B & Decoder & Instr. tuning & Last & No & 4096 & \href{https://huggingface.co/meta-llama/Meta-Llama-3-8B}{HF} \\
\bottomrule
\end{tabularx}
\caption{Overview of data representations used in our experiments, including architecture, pretraining objective, pooling strategy, chemistry adaptation, and embedding dimensionality and direct link to the source.}
\label{tab:repr-comparison}
\end{table*}
\footnotetext[1]{\textsuperscript{$\ast$} While OpenAI Embedding model architecture is not publicly disclosed, we assign a decoder-style structure based on the GPT family.}
\footnotetext[2]{\textsuperscript{$\dagger$} Partially chemistry-aligned through prompting instructions like "Represent the chemical reaction."}

\section{Benchmarking datasets}
\label{section:benchmark-datasets}
\subsection{Synthetic organic chemistry}
\paragraph{Buchwald-Hartwig reactions}
We performed all initial investigations on a set of Buchwald-Hartwig (BH) reactions with the task of optimizing yield (0-100\%). This dataset consists of 3955 reactions spanning across five distinct products (BH1-BH5). The data originates from a high-throughput experimentation (HTE) study published by Ahneman et al. \cite{ahneman2018predicting}. For each product, reactions were evaluated based on their percentage yield as determined by HPLC analysis. The design space is combinatorial across 15 reactants (aryl halides), 22 additives, 4 ligands, and 3 bases in DMSO solvent. 

In all initial BO experiments (unless explicitly stated as in T5Chem-SMILES), we represented reactions through procedural text template describing the reaction conditions in natural language. Moving forward to benchmarking against other models, we featurized reactions based on reaction SMILES, to ensure fair comparison to models that report best performance when using this representation (LAPEFT). Moreover, this dual representation allowed us to evaluate and establish robustness to the impact of different input formats on model performance. 

\paragraph{Additive screening}
This dataset originates from a study on organic additives' influence on the reactivity of complex Ni-catalysed reactions in a high-throughput experimentation (HTE) setup \cite{prieto2022accelerating}. It covers a wide range of screened additives (720) across four different reactions (Additives 1-4) and measures their effect on UV210 product area absorption. The challenge with traditional featurization methods in this dataset lies in the sole variability of the additive in the design space, while the other reaction parameters remain fixed. In such a setup, traditional one-hot encoding techniques would yield results similar to random search. Previous approaches report success with DRFP representation, comparing its performance to a set of molecular descriptors \cite{rankovic2023bayesian}. Compared to this method, we achieve better results while representing the reaction SMILES through LLM embeddings and optimizing with our GP-guided finetuning approach.

\paragraph{Suzuki-Miyaura reactions}
Suzuki-Miyaura dataset is a reaction optimization benchmark that includes high-throughput evaluated Suzuki-Miyaura cross-coupling reactions, originally introduced by Perera et al. \cite{perera2018platform}. The dataset was generated using an automated nanomole-scale synthesis platform designed to explore large combinatorial reaction spaces efficiently. It contains 5760 reactions across varied combinations of 7 unique electrophile, 4 nucleophile, 11 ligands (plus one blank), 7 bases (plus one blank), 4 solvents and Pd(OAc)2 as precatalyst, while the optimization objective is maximizing yield.

\subsection{Molecular optimization}
We selected the benchmark datasets from Kristiadi et al. \cite{kristiadi2024sober} to both test our model on molecular property optimization and compare directly to their approach (LAPEFT). Moreover, we use molecular SMILES as the textual representation of the data following their best practice and offering a fair comparison between the methods. The five datasets we present in this work span diverse scientific applications and optimization objectives:

\begin{itemize}
    \item Redox (1,407 samples): minimize redox potential for flow battery materials \cite{Agarwal2021},
    \item Solvation (1,407): minimize solvation energy \cite{Agarwal2021},
    \item Kinase (10,449): minimize docking score in kinase inhibitors \cite{Graff2021},
    \item Photoswitch (392): maximize the $\pi - \pi^*$ transition wavelength* in organic photoswitches \cite{Griffiths2022a},
    \item PCE (10,000): maximize power conversion efficiency of photovoltaic materials \cite{Lopez2016}.
\end{itemize}

All objectives are continuous and we use molecular fingerprints as a chemistry-related featurization for the baseline GP comparison. 

\subsection{Material science and catalysis}

To move beyond reaction and molecular optimization we include two datasets from Olympus \cite{hase_olympus_2021}, involving catalysis and crystallization, together with C2 yield optimization from \cite{ramos2023bayesian}, all of which involve continuous or mixed-variable optimization objectives. These datasets are commonly used in autonomous discovery and closed-loop optimization studies. 

\paragraph{Catalyst optimization -- C2 yield}

We evaluate our method on optimizing methane oxidative coupling (OCM) using a subset of 1180 reactions from \cite{ramos2023bayesian}. Each reaction involves synthesizing a supported catalyst (e.g., \ce{Mn-Na2WO4/BN}
) by impregnating a solid support (typically BN) with a solution of up to three metal precursors in defined molar ratios. Additional parameters include reaction temperatures (typically \textasciitilde
\SI{900}{\degreeCelsius}), with controlled gas flows (\ce{CH4}, \ce{O2}, Ar) and contact times. The objective is to maximize C2 yield, a measure of desirable product formation. The original study already provides a textual template of reactions in this dataset, which we used for testing our method on diverse input formatting. For the standard GP baseline we featurize the data by concatenating numerical parameters and one-hot encoded categorical values.

\paragraph{OER (Oxygen evolution reaction catalysts).}
This dataset comprises 2,121 samples describing compositions of high-throughput screened catalysts for the oxygen evolution reaction. Each data point represents a combination of elemental loadings (Ni, Fe, Co, Mn, Ce, La) constrained to sum to 1. The optimization goal is to minimize the overpotential, a key descriptor of catalytic efficiency.

\begin{itemize}
    \item \textbf{Target:} Overpotential (continuous)
    \item \textbf{Features:} 6 discrete fractional loadings
\end{itemize}

\vspace{1em}

\paragraph{Vapdiff crystallization (crystal score).}
This dataset reports the outcomes of vapor diffusion crystallization experiments across 918 combinations of organic, solvent, and inorganic conditions. The target is an \emph{ordinal} score representing crystallization quality, with categorical and continuous inputs describing the experiment setup.

\begin{itemize}
    \item \textbf{Target:} Crystal score (ordinal)
    \item \textbf{Features:} 10 variables (categorical, continuous, discrete)
\end{itemize}
Similarly to C2 yield optimization, we featurize the categorical variables through one-hot encoding and concatenate these vectors to the remaining numerical parameters for the input to the standard GP baseline.

\vspace{1em}

\subsection{Analytic and process chemistry}

\paragraph{HPLC (High-Performance Liquid Chromatography).}
This dataset includes 1,386 data points measuring peak response from an automated HPLC system as a function of six continuous process parameters such as flow rate, sample volume, and wait time. The objective is to maximize the peak signal (measured by photo degradation response).

\begin{itemize}
    \item \textbf{Target:} Photo degradation (continuous)
    \item \textbf{Features:} 6 continuous parameters
\end{itemize}

\clearpage
\section{Extended results}

\subsection{Tokenization per LLM type}

\begin{wrapfigure}{r}{0.45\textwidth}
\centering
\vspace{-15pt}
\includegraphics[width=\linewidth]{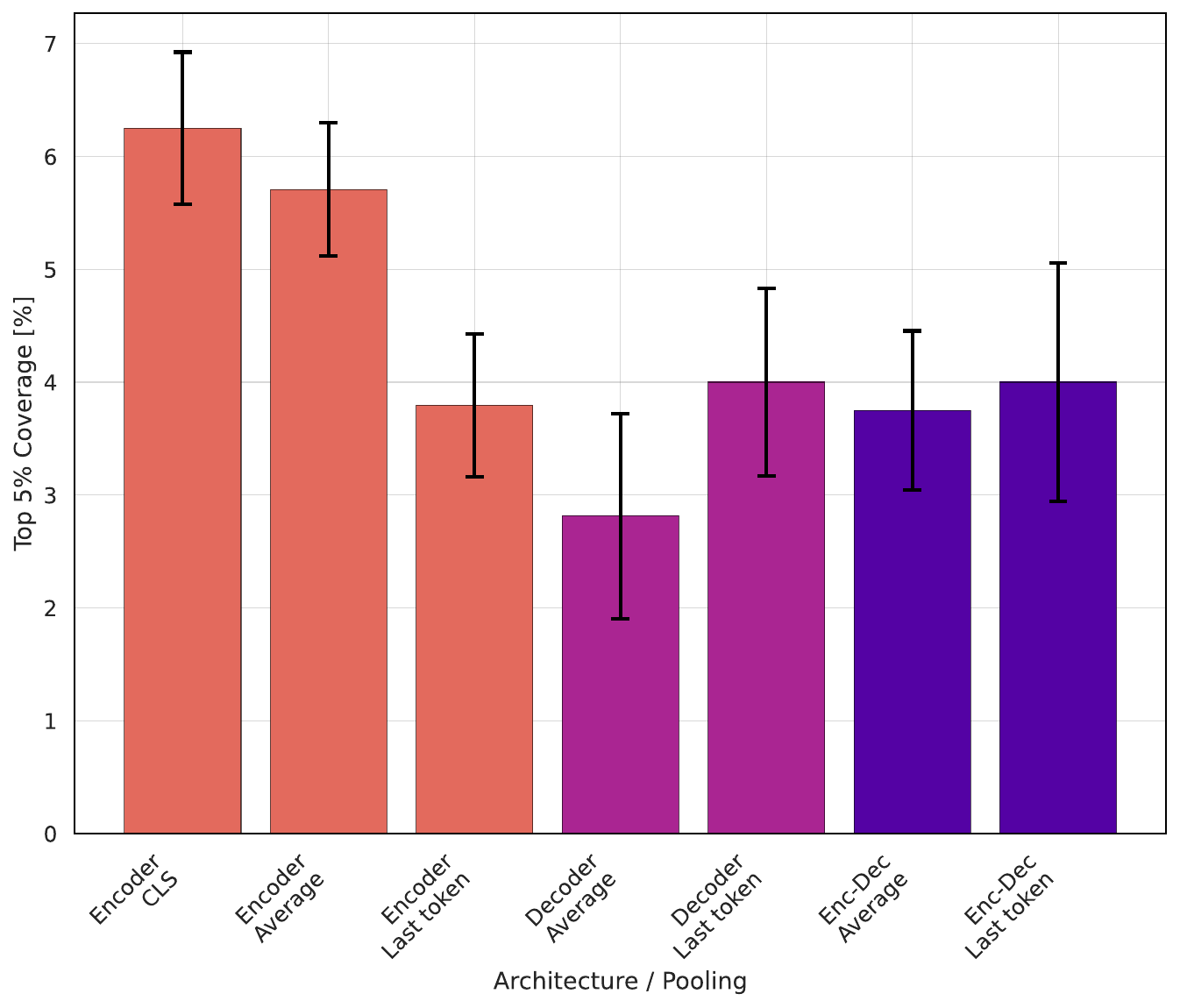}
\caption{\textbf{Tokenization pool strategy per LLM type.} We compare the pooling strategies across LLM types. Encoder-based models benefit from utilizing the CLS token, unlike decoder architectures where this token collapses all inputs to duplicated representation (hence not appearing in the results). For decoder-based architectures, last token pooling improves results over token averaging. For encoder-decoder models, the difference between average and last token pool is less pronounced, however with a lower variability for mean pooling. The bars represent the standard error and we compare results on BH1 reaction with 25 BO iterations repeating the experiments over 10 seeds. }
\label{fig:tokenization}
\end{wrapfigure}

We investigate the impact of different pooling strategies on the quality of LLM embeddings for Bayesian optimization. Since LLMs produce variable-length token sequences, pooling plays a critical role in converting these sequences into fixed-size representations used by the GP surrogate. Our ablation reveals a clear interaction between model architecture and pooling choice. For encoder-only models, CLS token pooling outperforms alternatives -- last token pooling dilutes the informative signal captured in the CLS token -- specifically trained to represent global context. In contrast, decoder-only models tend to collapse all inputs to similar representations when the starting token is pooled, leading to duplicates and unsuccessful optimization. Here, last-token pooling aligns with the autoregressive structure and yields substantially better results. For encoder-decoder models, we observe lower differences in performance across pooling strategies, though mean pooling shows a slight edge in consistency.

Based on these findings, we adopt CLS pooling for encoder models, last-token pooling for decoder models, and mean pooling for encoder-decoder models throughout the main experiments. Figure~\ref{fig:tokenization} summarizes the performance differences across model types and pooling methods. The pooling choices ensure meaningful input representations across LLM types, and deviations from optimal setup can lead to noticeable performance drops within the fixed-feature setting (e.g., up to 40\% in top-5 discovery rate between CLS and last-token pool for encoder models).

\subsection{Which LLM layers carry the most information?}

\begin{figure}[h!]
\begin{center}
\includegraphics[width=0.99\linewidth]{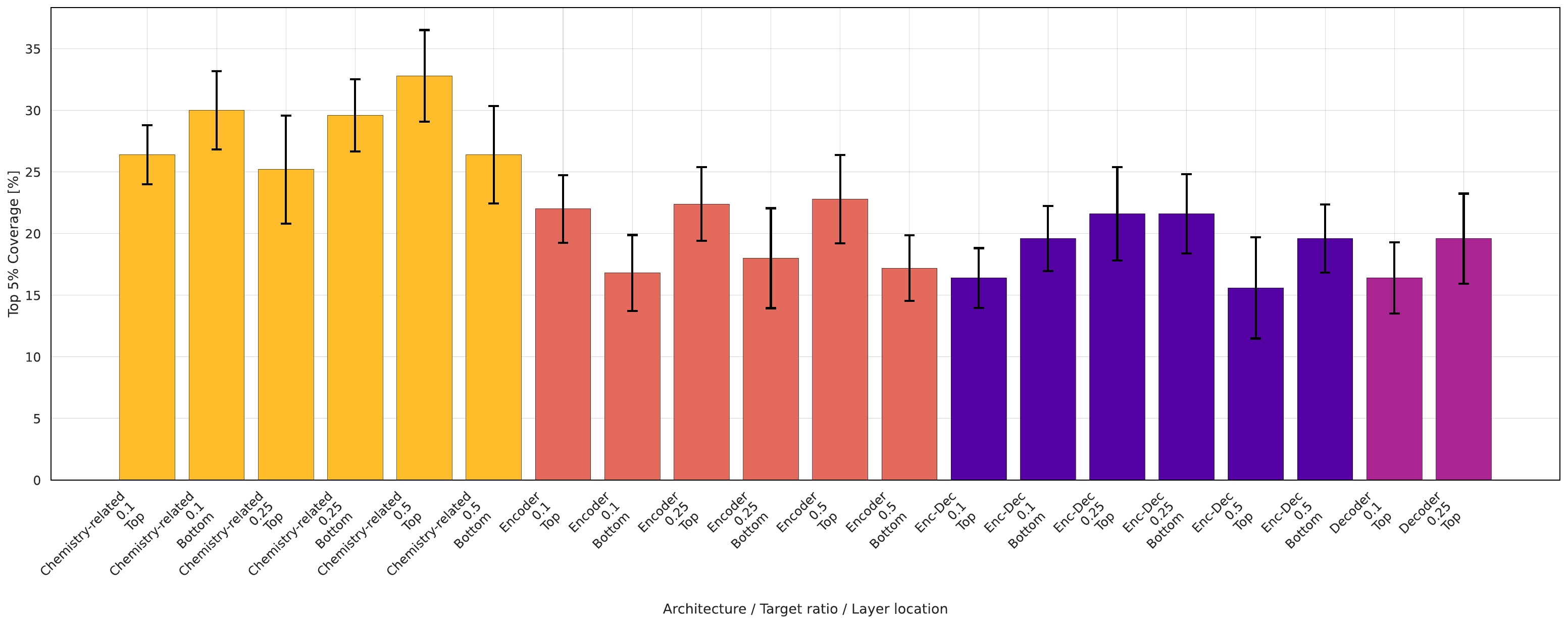}
\end{center}
\caption{\textbf{Performance comparison of PEFT strategies across LLM architectures.} We vary the proportion (10\%, 25\%, 50\%) and location (top vs. bottom) of targeted linear layers using LoRA. Results show that top-layer finetuning consistently outperforms bottom-layer updates for encoder-only and decoder-only{$\ast$} models. For encoder-decoder models, performance is more consistent across layer locations. Based on these findings, we fix the default to targeting the top 25\% of linear layers.}
\label{fig:layers}
\end{figure}

\begin{figure}[h!]
\begin{center}
\includegraphics[width=\linewidth]{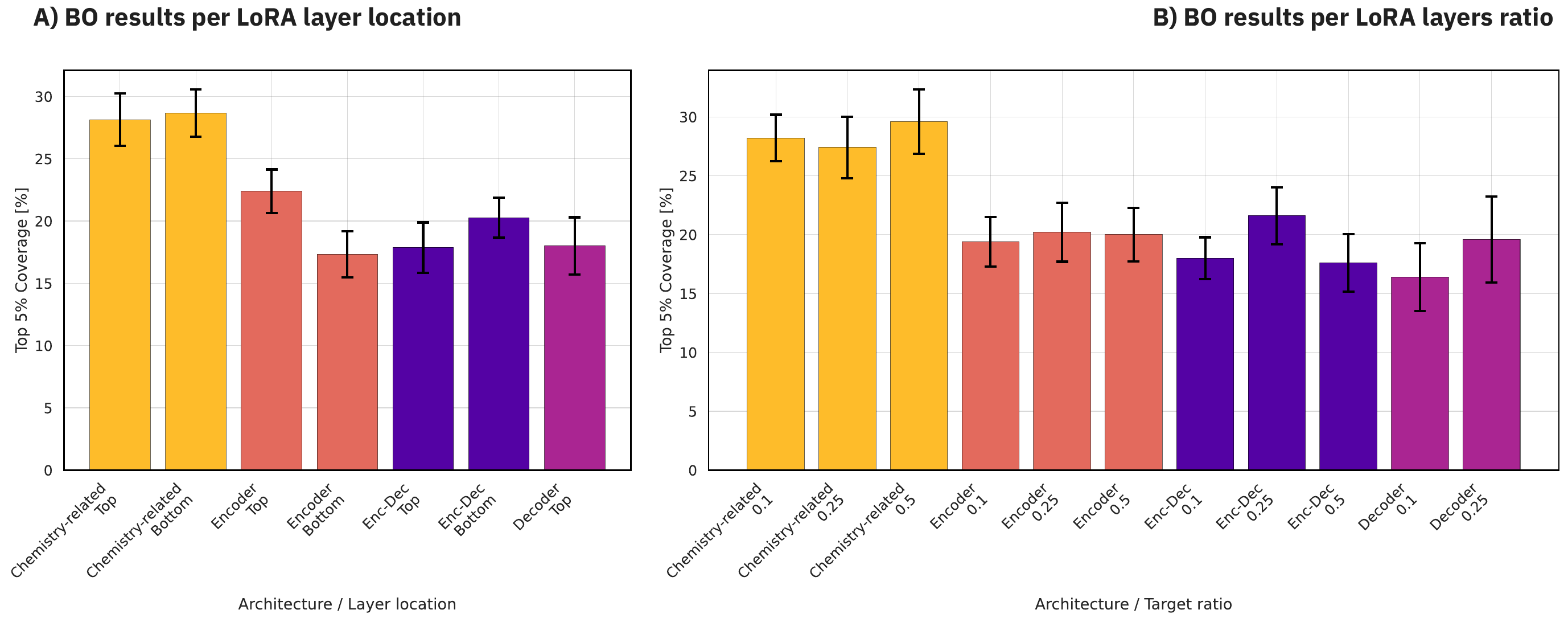}
\end{center}
\caption{Breakdown of results from Figure \ref{fig:layers}, highlighting the effect of LoRA layer location (top vs. bottom) and proportion (10\%, 25\%, 50\%) on BO performance. Targeting bottom layers in decoder-only models (e.g., Qwen2-7B) resulted in numerical instabilities during optimization while higher proportion (50\%) resulted in out-of-memory issues.}
\label{fig:layers_per_ratio_per_location}
\end{figure}

We further examine which subset of LLM layers is most effective to target during PEFT. Since full model finetuning within the GP kernel is infeasible for large LLMs, we use LoRA to adapt only a fraction of the model weights. The choice of which layers to modify has a strong effect on optimization performance. Inspired by the intuition that higher (deeper) transformer layers encode more task-relevant semantics, we evaluate LoRA targeting strategies by varying both the location (top vs. bottom layers) and proportion (10\%, 25\%, 50\%) of modified linear layers.

Figures~\ref{fig:layers} and ~\ref{fig:layers_per_ratio_per_location} show the results across four (we now separate the encoder-decoder models into chemistry-related--T5Chem and base T5) LLM architectures. For encoder-only, targeting the top layers consistently outperforms bottom-layer finetuning. This aligns with the well-established view that deeper layers in such models encode more abstract and domain-specific representations. Importantly, we omit bottom-layer results for decoder-only models (e.g., Qwen2-7B) due to instability and numerical issues encountered during training, likely caused by incompatible LoRA insertions in early layers, and only show results of the different ratio of LoRA adapted top layers (10\% and 25\%).

Interestingly, encoder-decoder models (both T5 and chemistry-specialized T5Chem) show competitive performance even when targeting bottom layers, though the top 25\% for T5 still yields slightly better or equally stable results. This suggests that meaningful information may be distributed across layers in encoder-decoder setups, potentially due to the dual role of encoding and decoding steps. For T5Chem, targeting the bottom 50\% of linear layers with LoRA yield the best results in the ablation (Figure \ref{fig:layers}). This observation could potentially justify targeting all layers with LoRA in chemistry-related architectures which could contribute to even better results. Nevertheless, to ensure a consistent and generalizable comparison across LLM sizes and types, we adopt a default strategy of targeting the top 25\% of linear layers for all models. This decision achieves a balance between performance and computational efficiency, while also avoiding the overhead of architecture-specific tuning.

\footnotetext[1]{\textsuperscript{$\ast$} Targeting bottom layers in decoder-only models (Qwen2-7B) led to numerical instabilities in optimization.}

\subsection{Structure vs fit and BO performance}

Selecting suitable priors is one of the core challenges in BO, especially when limited information is available about the underlying objective function. The first modeling choice in BO is how to represent the input design space. In chemistry, this space can be expressed in various ways, such as SMILES strings, reaction templates, or molecular fingerprints. Often, this representation is predetermined by the constraints of the problem setting, such as one-hot encodings in combinatorial screens. However, the choice of representation imposes downstream consequences on other components of the BO pipeline.

Surrogate models, which map the input $\mathbf{x}$ to the output $y$ in a probabilistic manner, come with their own inductive biases. GPs rely on a kernel function to define similarity between points. The choice of kernel encodes assumptions about smoothness, differentiability, and the geometry of the function to be modeled. For example, the Tanimoto kernel \cite{griffiths2023gauche} might be better suited for binary fingerprint inputs, while Matérn kernels are broadly applicable to continuous Euclidean representations. In our study, we fix the surrogate kernel to Matérn 5/2 due to its balance between smoothness and flexibility, and its prevalence in chemical BO applications. All related analysis in this section is, therefore, built on the basis of this kernel. Future work may explore kernel-specific behaviors in structure/fit alignment. With the selected design space representation and the surrogate kernel, we are left with a set of assumptions about the structure of the objective function itself. If these assumptions are misspecified -- e.g., if the representation induces a geometry not aligned with the kernel -- then BO performance can degrade.

We now analyze how the choice of data representation impacts BO performance under a fixed surrogate model and acquisition function. To do so, we introduce a single normalized smoothness metric: the ratio between the GP's learned lengthscale and the average pairwise distance in the embedding space. This metric reflects how far the GP generalizes relative to the data distribution, serving as a proxy for the compatibility between the representation space and the kernel's inductive bias.

We observe a strong correlation ($r = 0.92$) between this normalized lengthscale and BO performance (Figure \ref{fig:bh-results}c), suggesting that representations that allow the GP to maintain broader, smoother fits tend to support more successful optimization. A higher ratio indicates that the GP can generalize over broader regions while still resolving performance differences, ultimately leading to better acquisition decisions. This trend is consistent with the intuition that smoother fits -- enabled by coherent, well-structured representation spaces -- support more principled exploration and reduce overfitting to local noise.

\begin{figure}
    \centering
    \includegraphics[width=\linewidth]{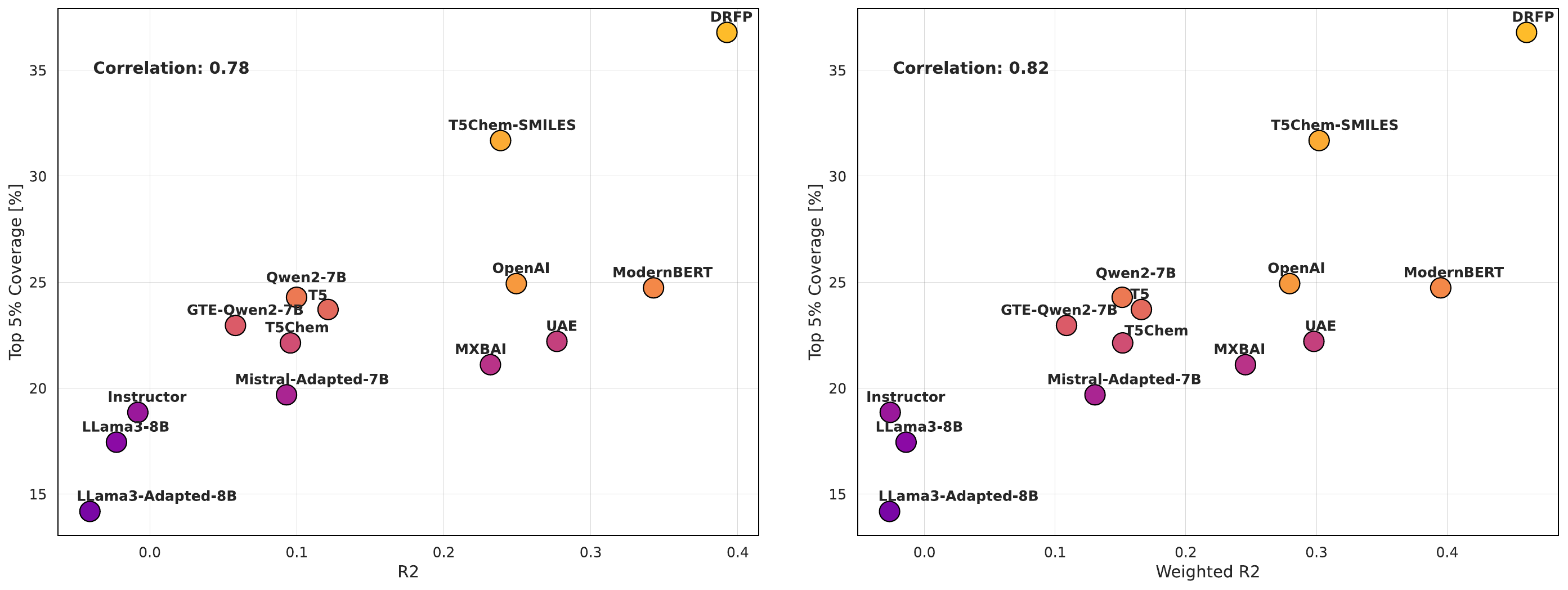}
    \caption{We compare (left) $R^2$ over the entire design space, (right) weighted $R^2$ that upweights points in the top 5\% (using a 3:1 weighting scheme). While fit alone is informative ($r = 0.78$ unweighted, $r = 0.82$ weighted), the smoothness metric based on normalized lengthscale achieves a stronger correlation with BO success ($r = 0.92$, main text Figure~\ref{fig:bh-results}), highlighting the importance of representation-structure alignment.}
    \label{fig:struc-vs-fit-corr}
\end{figure}

We further show that while standard and weighted $R^2$ measures correlate with BO performance, their predictive power is consistently lower than the normalized smoothness metric. This supports the view that while accurate fit helps, smooth fits -- enabled by representations that align well with the GP kernel -- are even more important for effective acquisition. This finding motivates our proposed deep kernel learning approach. By jointly training the LLM and GP via marginal likelihood, we allow the representation to adapt to the GP’s inductive assumptions, resulting in smoother surrogate fits and more structured latent spaces -- as shown in Figure \ref{fig:gp-fit-dist-tsne}. This ultimately enables better optimization.

We also observe that the best-performing fixed-feature baseline (DRFP) already exhibits relatively structured embedding space, reflected in clear pairwise L2 and kernel similarity histograms. In contrast, the fixed T5 model produces less organized latent structure. With our adaptive method (PLLM$\phi$+T5), however, the latent space becomes substantially more structured, resulting in smoother GP fits and better-aligned similarity distributions. This confirms the key role of aligning learned representations with the GP kernel’s inductive bias.

\label{sec:bo-fit-analysis}

\begin{figure}
    \centering
    \includegraphics[width=\linewidth]{figures/supplementary/gp-fit-tsne-distances.pdf}
    \caption{\textbf{Visual analysis of latent space structure, GP behavior, and similarity metrics across different representations.}
    We compare three models: (A) GP+DRFP, the best-performing fixed-feature baseline; (B) BoChemian+T5, using frozen LLM embeddings from natural language templates; and (C) PLLM$\phi$+T5, our best adaptive embedding model. The left two columns visualize the latent space with ground truth (left) and GP posterior mean/variance (middle) colored by yield. We mark the suggested and initial points. The right two columns show pairwise kernel similarities and L2 distances for high--high, high--low, and low--low yielding regions. DRFP exhibits mild structural organization even in its fixed feature space, contributing to strong performance. T5 without finetuning lacks this structure, while PLLM$\phi$+T5 learns a highly structured latent space, enabling smoother fits and more effective acquisition decisions.}
    \label{fig:gp-fit-dist-tsne}
\end{figure}

\subsection{BO results per representation type}

\begin{figure}
\begin{center}
\includegraphics[width=0.99\linewidth]{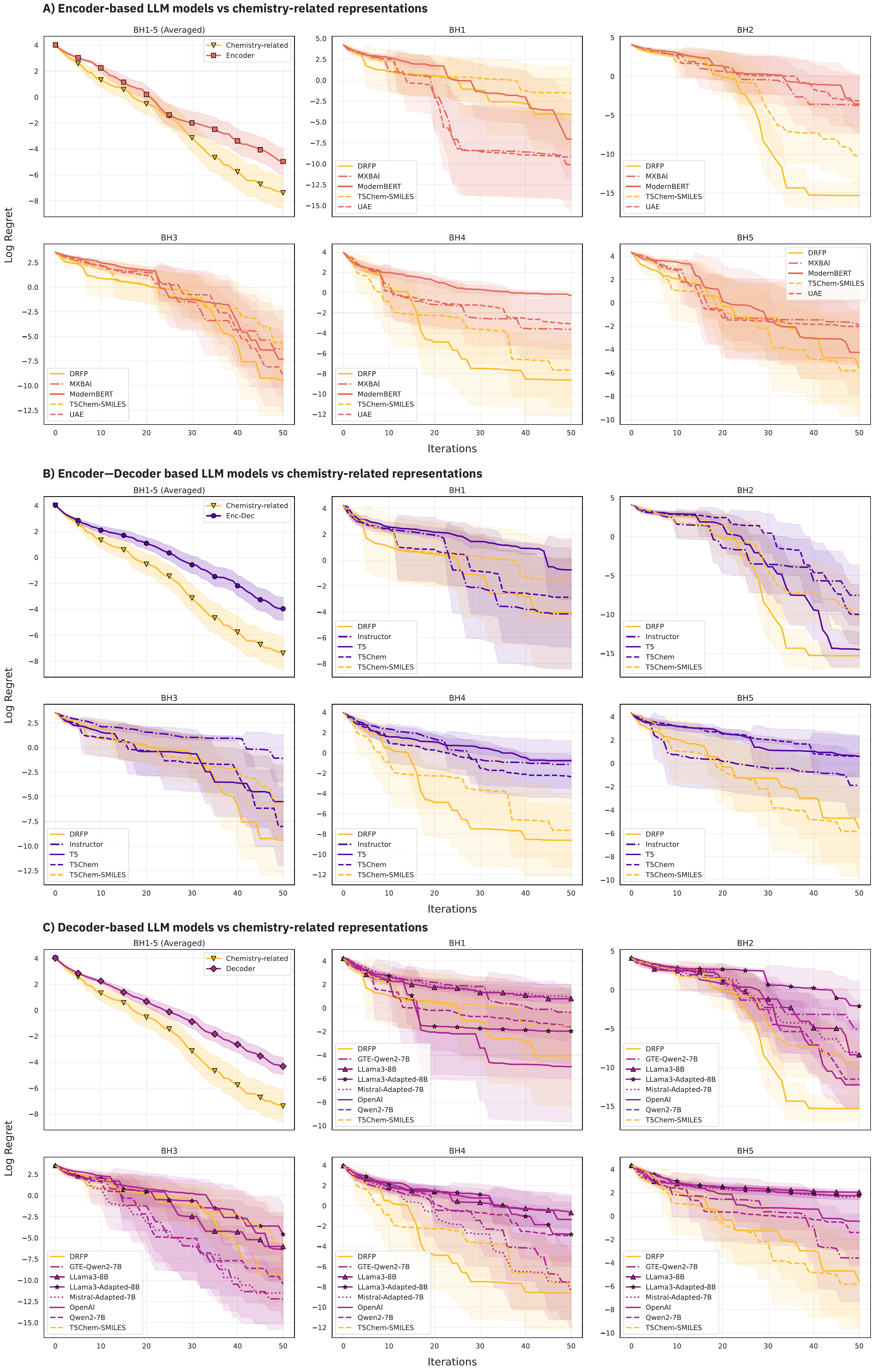}
\end{center}
\caption{Individual LLM BO tracelines in Buchwald-Hartwig optimization.}
\label{fig:bo-traces-individual}
\end{figure}

With the BO traces in Figure \ref{fig:bo-tracelines} we show aggregated results per LLM or chemistry-related representation types. Additionally, we provide an overview of the BO performance for each individual representation (LLM or chemistry related) during the 50 optimization steps in Figure \ref{fig:bo-traces-individual}. We observe that performance varies in different BH reactions, with no representation consistently outperforming others across all tasks -- including chemistry-specialized ones. All LLM types, however, show similar distributions of suggested point evaluations. In comparison to other LLM architectures, encoder-based models tend to achieve higher $R^2$ values during optimization. However, the improved function approximation does not necessarily translate to better BO performance, as modeling the function and identifying its optimal points are two fundamentally distinct, though complementary, objectives.

\begin{figure}[h!]
\begin{center}
\includegraphics[width=\linewidth]{figures/supplementary/bo-traces.pdf}
\end{center}
\caption{\textbf{BO metrics per LLM type.} A) We show optimization paths for all BH reactions (BH1-5 Averaged) across different LLM types (Encoder only, Encoder-Decoder, Decoder only) and chemistry-related representations (DRFP, T5Chem-SMILES) together with optimization results on individual reactions (BH1-BH5). B) Distribution of evaluated suggestions generated throughout the entire optimization process (50 iterations) for 20 seed runs and all BH reactions. C) $R^2$ scores per LLM type over the evolving design space, averaged across all BH reactions.}
\label{fig:bo-tracelines}
\end{figure}

\begin{figure}
\begin{center}
\includegraphics[width=0.99\linewidth]{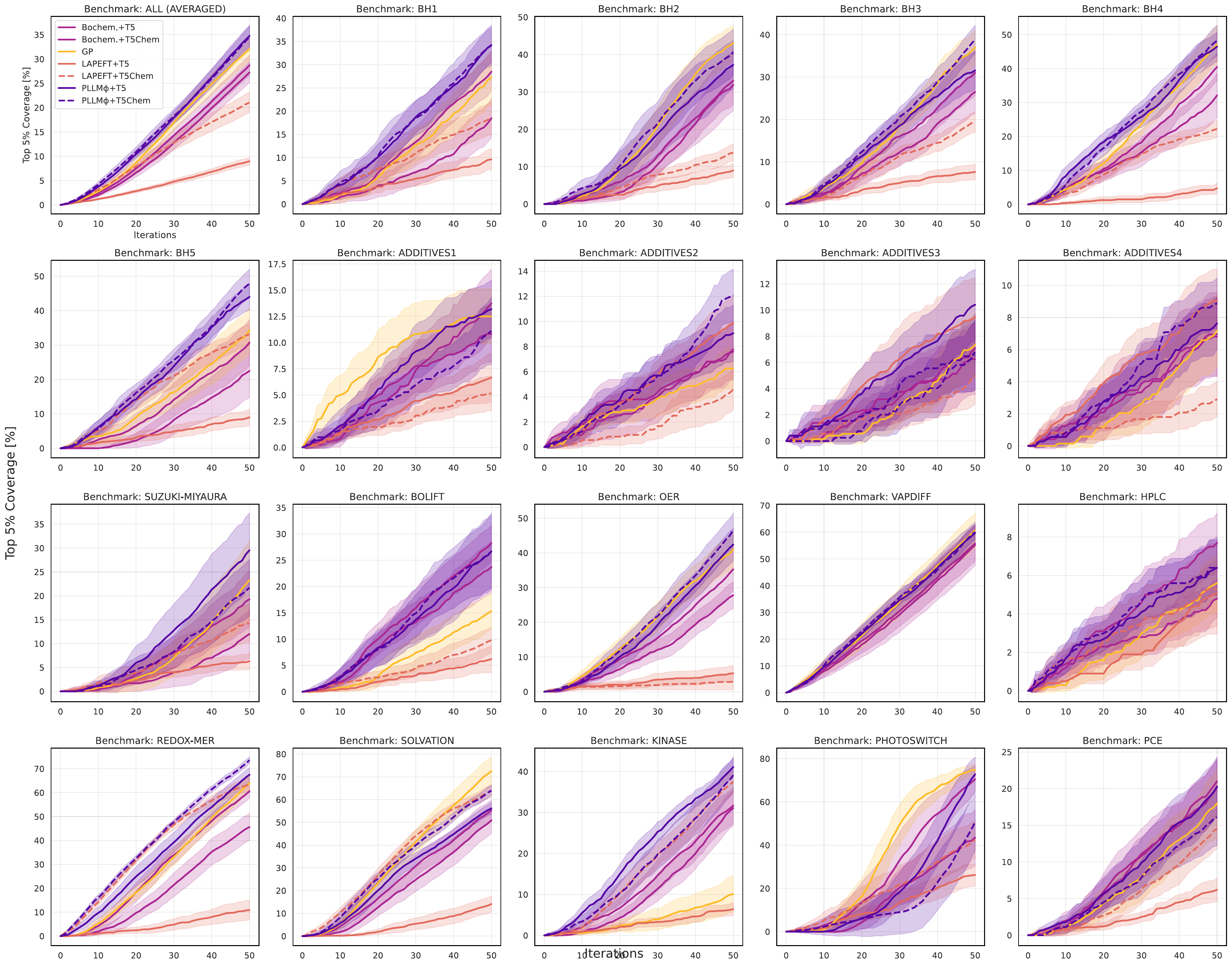}
\end{center}
\caption{ Top 5\% coverage per iteration for all benchmarks.}
\label{fig:all-benchmarks-accros-iter}
\end{figure}

\clearpage
\section{Reproducibility}
\label{appendix_reproducibility}
\subsection{BO Initialization}

\begin{wrapfigure}{r}{0.42\textwidth}
\centering
\vspace{-10pt}
\includegraphics[width=\linewidth]{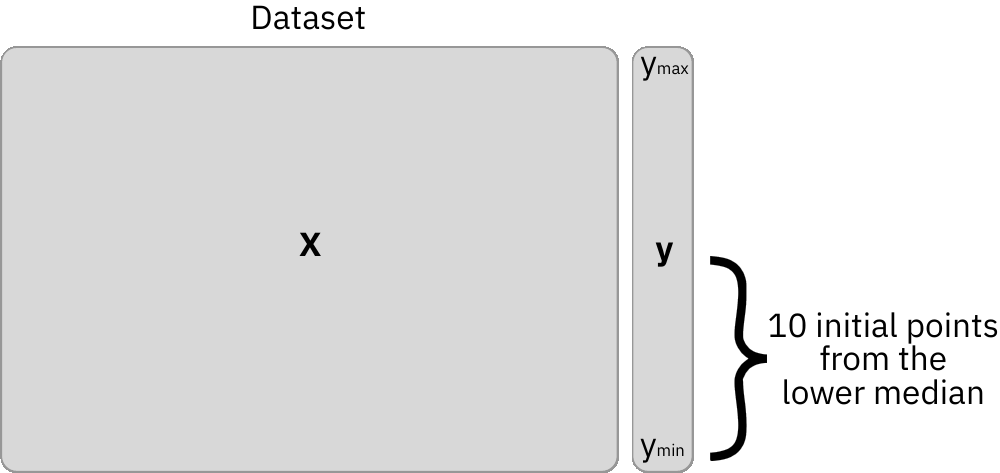}
\caption{\textbf{Illustrative example of initial data selection.} For all benchmark datasets, we constrain the selection of initial points for BO to the lower median points based on the objective values. This initialization strategy better reflects real-world scenarios where optimization starts from suboptimal conditions. }
\label{fig:bo-init}
\end{wrapfigure}

Bayesian optimization is in practice initialized with a small number of pre-existing datapoints, either from prior experimentation or simulation. In reaction optimization, for example, these initial points often correspond to unsuccessful or low-yielding reactions. In scientific discovery settings, this setup is not only realistic but expected -- optimization typically begins from sparse and suboptimal conditions, with the goal of efficiently identifying high-performing regions.

This starting point contrasts with scenarios where good conditions are already known, in which case optimization reduces to local exploitation rather than global search. Reaching high-yielding conditions from poor initial data is substantially more challenging and better reflects real-world discovery pipelines. Compounding this difficulty is the negative bias in the scientific literature: failed or low-yield experiments are rarely reported, making published datasets inherently skewed toward successful outcomes \cite{ioannidis2005most}. 

To simulate this setting, we initialize the BO algorithm (Algorithm~\ref{alg:bo}) with 10 points sampled from the lower median range of the objective value distribution. Figure~\ref{fig:bo-init} illustrates this strategy. For each benchmark dataset, the candidate pool is first sorted by objective value, and initial points are randomly selected from those with values below the median. This initialization requires the model to reason under uncertainty and efficiently navigate toward high-performing regions with minimal prior knowledge.

\subsection{BO setup}
Following the initialization, we run the BO loop for 50 iterations with batch size 1. We repeat each experiment configuration for 20 times with different seed values (1-20) to obtain robust performance metrics. Our choice for the GP kernel is Matérn-5/2, based on its demonstrated effectiveness on both continuous and discrete design spaces \cite{rankovic2023bayesian}. For balancing the exploration and exploitation we employ the expected improvement acquisition function.

\subsection{Implementation details}
\label{sup:implementation}

\paragraph{Surrogate Models}
We implement both fixed-feature and finetuned surrogate models as subclasses of \texttt{SingleTaskGP} from \texttt{botorch} \cite{balandat2020botorch}. For fixed-feature GPs, the inputs are LLM embeddings, chemistry-related representations (fingerprints, DRFP) or default parameters (one-hot encoded categorical variables, numerical values), while we learn the GP kernel hyperparameters \(\theta = \{\ell, \sigma^2, \sigma_n^2, c\}\) by maximizing the marginal likelihood using the L-BFGS-B optimizer provided by BoTorch's \texttt{fit\_gpytorch\_mll} routine. We employ the Matérn-5/2 kernel \cite{gardner2018gpytorch} with initialization: \(\ell=1.0\), \(\sigma^2=1.0\), and \(\sigma_n^2=1.0 \times 10^{-4}\). The optimizer runs with multiple restarts, as part of BoTorch's default behavior.

For deep kernel GPs, we extend the surrogate to jointly optimize both GP and LLM parameters. We build a custom \texttt{DeepGP} class that incorporates a finetuning model (PEFT adapter and/or projection head) jointly trained via AdamW \cite{loshchilov2017decoupled}. We optimize GP and LLM parameters using separate learning rates (\(2 \times 10^{-1}\) for GP, \(2 \times 10^{-3}\) for LLM) with a shared weight decay of \(1 \times 10^{-3}\). We apply gradient clipping with a max norm of 1.0 and decay the learning rates using a StepLR scheduler with a decay factor of 0.95.

\paragraph{PEFT Configuration}
We insert LoRA adapters into the top 25\% of linear layers, using the following configuration: rank \(r=4\), \(\alpha=16\), no bias updates and dropout of 0.2. 

\paragraph{Projection Layer}
We define the projection layer as:
\begin{equation}
\mathbf{z} = \text{ELU}(\text{Dropout}(W \mathbf{x} + b)),
\end{equation}
where \(W \in \mathbb{R}^{d \times 64}\) and input \(d\) is model-dependent. We initialize the parameters of this model via Xavier uniform initialization.

\paragraph{Featurization}
We extract LLM embeddings from Hugging Face models using their default tokenizers and truncating the input to maximum of 512 tokens. Pooling strategies depend on the model architecture: CLS token for encoder-based models, last-token for decoder-only models, and mean-pooling for encoder-decoder models. 

\paragraph{Training setup}
Fixed-feature models can be run on CPU or a single GPU (e.g., RTX 3090 with 24GB VRAM). Finetuned experiments involving large LLMs (e.g., PLLM$\phi$+Qwen2-7B) are run on NVIDIA H100 GPU and 96GB RAM. Lightweight models (e.g., PLLM$\phi$+T5, PLLM$\phi$+ModernBERT) are trainable on local hardware, with under 300k trainable parameters in total (e.g., 230k out of 149M for ModernBERT, 165k out of 109M for T5; less than 0.2\% of model weights updated).

\paragraph{Tracking and seeding}
All experiments are seeded using \texttt{seed\_everything} from \texttt{pytorch\_lighting} \cite{Falcon_PyTorch_Lightning_2019} with values 1-20. Acquisition optimization is deterministic over the candidate pool. We do not augment the data or use any stochastic featurization. We use \texttt{wandb} \cite{wandb} for tracking experiments, managing seeds, logging losses, metrics, learning rates, images, and running configuration sweeps.

\section{Supporting tables}

\renewcommand{\arraystretch}{1.2}
\begin{longtable}{lc}
\hline
Representation & Top 5\% Coverage [\%] \\
\hline
\endfirsthead
\hline
Representation & Top 5\% Coverage [\%] \\
\hline
\endhead
DRFP & \textbf{37.750 $\pm$ 13.983} \\
\hline
\nopagebreak
T5Chem-SMILES & 32.750 $\pm$ 13.728 \\
\hline
\nopagebreak
OpenAI & 25.650 $\pm$ 12.287 \\
\hline
\nopagebreak
ModernBERT & 25.475 $\pm$ 10.061 \\
\hline
\nopagebreak
Qwen2-7B & 25.150 $\pm$ 13.934 \\
\hline
\nopagebreak
T5 & 24.450 $\pm$ 13.005 \\
\hline
\nopagebreak
GTE-Qwen2-7B & 23.500 $\pm$ 15.288 \\
\hline
\nopagebreak
T5Chem & 22.950 $\pm$ 11.571 \\
\hline
\nopagebreak
UAE & 22.800 $\pm$ 13.144 \\
\hline
\nopagebreak
MXBAI & 21.800 $\pm$ 14.733 \\
\hline
\nopagebreak
Mistral-Adapted-7B & 20.350 $\pm$ 13.046 \\
\hline
\nopagebreak
Instructor & 19.300 $\pm$ 10.935 \\
\hline
\nopagebreak
LLama3-8B & 18.056 $\pm$ 12.745 \\
\hline
\nopagebreak
LLama3-Adapted-8B & 14.625 $\pm$ 10.803 \\
\hline
\nopagebreak
Random Search & 6.176 $\pm$ 3.751 \\
\hline
\caption{BO with fixed LLM features. }
\end{longtable}

\renewcommand{\arraystretch}{1.2}
\begin{longtable}{lc}
\hline
Representation & Top 5\% Coverage [\%] \\
\hline
\endfirsthead
\hline
Representation & Top 5\% Coverage [\%] \\
\hline
\endhead
PLLM$\phi$+T5 & \textbf{42.602 $\pm$ 13.111} \\
\hline
\nopagebreak
LLM$\phi$+T5 & 42.577 $\pm$ 13.817 \\
\hline
\nopagebreak
PLLM+T5 & 41.075 $\pm$ 10.844 \\
\hline
\nopagebreak
PLLM$\phi$+Qwen2-7B & 40.250 $\pm$ 11.189 \\
\hline
\nopagebreak
PLLM+Qwen2-7B & 39.375 $\pm$ 9.850 \\
\hline
\nopagebreak
LLM$\phi$+Qwen2-7B & 38.725 $\pm$ 12.780 \\
\hline
\nopagebreak
PLLM$\phi$+ModernBERT & 36.250 $\pm$ 14.100 \\
\hline
\nopagebreak
LLM$\phi$+ModernBERT & 35.775 $\pm$ 13.618 \\
\hline
\nopagebreak
PLLM+ModernBERT & 34.600 $\pm$ 10.805 \\
\hline
\nopagebreak
PLLM+OpenAI & 33.525 $\pm$ 11.791 \\
\hline
\nopagebreak
OpenAI & 25.650 $\pm$ 12.287 \\
\hline
\nopagebreak
ModernBERT & 25.475 $\pm$ 10.061 \\
\hline
\nopagebreak
Qwen2-7B & 25.150 $\pm$ 13.934 \\
\hline
\nopagebreak
T5 & 24.450 $\pm$ 13.005 \\
\hline
\caption{BO results with LLM-based deep kernels. }
\end{longtable}

\renewcommand{\arraystretch}{1.2}
\begin{longtable}{llcc}
\hline
Benchmark & Method & $R^2 \uparrow$ & $NLPD \downarrow$ \\
\hline
\endfirsthead
\hline
Benchmark & Method & $R^2 \uparrow$ & $NLPD \downarrow$ \\
\hline
\endhead
\multirow{7}{*}{ADDITIVES-1} & PLLM$\phi$+T5 (ours) & -0.00 $\pm$ 0.10 & 21.19 $\pm$ 2.77 \\
 & PLLM${\phi}$+T5Chem (ours) & 0.07 $\pm$ 0.09 & 27.02 $\pm$ 3.93 \\
 & GP+DRFP & 0.09 $\pm$ 0.06 & 11.34 $\pm$ 0.11 \\
 & Bochem.+T5 & 0.14 $\pm$ 0.05 & 11.66 $\pm$ 0.92 \\
 & Bochem.+T5Chem & 0.17 $\pm$ 0.09 & 11.87 $\pm$ 1.25 \\
 & LAPEFT+T5 & -0.15 $\pm$ 0.14 & 723.23 $\pm$ 1625.77 \\
 & LAPEFT+T5Chem & -0.04 $\pm$ 0.13 & 329.75 $\pm$ 566.60 \\
\hline
\nopagebreak
\multirow{7}{*}{ADDITIVES-2} & PLLM$\phi$+T5 (ours) & -0.22 $\pm$ 0.13 & 25.72 $\pm$ 8.98 \\
 & PLLM$\phi$+T5Chem (ours) & -0.11 $\pm$ 0.09 & 36.73 $\pm$ 8.29 \\
 & GP+DRFP& 0.01 $\pm$ 0.02 & 10.00 $\pm$ 0.24 \\
 & Bochem.+T5 & 0.02 $\pm$ 0.03 & 10.58 $\pm$ 1.56 \\
 & Bochem.+T5Chem & 0.05 $\pm$ 0.03 & 10.21 $\pm$ 0.67 \\
 & LAPEFT+T5 & -0.15 $\pm$ 0.11 & 392.67 $\pm$ 516.44 \\
 & LAPEFT+T5Chem & -0.12 $\pm$ 0.10 & 292.60 $\pm$ 346.04 \\
\hline
\nopagebreak
\multirow{7}{*}{ADDITIVES-3} & PLLM$\phi$+T5 (ours) & -0.30 $\pm$ 0.31 & 33.75 $\pm$ 15.29 \\
 & PLLM$\phi$+T5Chem (ours) & -0.19 $\pm$ 0.21 & 52.34 $\pm$ 28.62 \\
 & GP+DRFP & -0.01 $\pm$ 0.02 & 10.36 $\pm$ 0.60 \\
 & Bochem.+T5 & -0.01 $\pm$ 0.02 & 10.50 $\pm$ 0.80 \\
 & Bochem.+T5Chem & -0.00 $\pm$ 0.02 & 10.48 $\pm$ 0.92 \\
 & LAPEFT+T5 & -0.21 $\pm$ 0.16 & 641.89 $\pm$ 1356.60 \\
 & LAPEFT+T5Chem & -0.12 $\pm$ 0.12 & 386.27 $\pm$ 431.61 \\
\hline
\nopagebreak
\multirow{7}{*}{ADDITIVES-4} & PLLM$\phi$+T5 (ours) & -0.34 $\pm$ 0.23 & 22.36 $\pm$ 4.87 \\
 & PLLM$\phi$+T5Chem (ours) & -0.21 $\pm$ 0.14 & 28.60 $\pm$ 9.03 \\
 & GP+DRFP & 0.01 $\pm$ 0.03 & 10.27 $\pm$ 0.13 \\
 & Bochem.+T5 & 0.01 $\pm$ 0.05 & 11.12 $\pm$ 2.58 \\
 & Bochem.+T5Chem & 0.00 $\pm$ 0.05 & 10.83 $\pm$ 1.41 \\
 & LAPEFT+T5 & -0.22 $\pm$ 0.15 & 186.94 $\pm$ 177.56 \\
 & LAPEFT+T5Chem & -0.14 $\pm$ 0.15 & 389.52 $\pm$ 715.90 \\
\hline
\nopagebreak
\multirow{7}{*}{BH-1} & PLLM$\phi$+T5 (ours) & 0.68 $\pm$ 0.05 & 5.57 $\pm$ 0.63 \\
 & PLLM$\phi$+T5Chem (ours) & 0.69 $\pm$ 0.07 & 7.11 $\pm$ 0.90 \\
 & GP+DRFP & 0.31 $\pm$ 0.34 & 6.14 $\pm$ 7.20 \\
 & Bochem.+T5 & 0.14 $\pm$ 0.23 & 19.96 $\pm$ 30.81 \\
 & Bochem.+T5Chem & 0.52 $\pm$ 0.14 & 32.23 $\pm$ 25.51 \\
 & LAPEFT+T5 & 0.51 $\pm$ 0.11 & 65.15 $\pm$ 83.74 \\
 & LAPEFT+T5Chem & 0.60 $\pm$ 0.04 & 216.41 $\pm$ 276.61 \\
\hline
\nopagebreak
\multirow{7}{*}{BH-2} & PLLM$\phi$+T5 (ours) & 0.65 $\pm$ 0.08 & 5.96 $\pm$ 1.00 \\
 & PLLM$\phi$+T5Chem (ours) & 0.65 $\pm$ 0.09 & 7.23 $\pm$ 1.17 \\
 & GP+DRFP & 0.50 $\pm$ 0.27 & 4.06 $\pm$ 0.77 \\
 & Bochem.+T5 & 0.22 $\pm$ 0.22 & 60.34 $\pm$ 102.47 \\
 & Bochem.+T5Chem & 0.50 $\pm$ 0.08 & 64.15 $\pm$ 58.29 \\
 & LAPEFT+T5 & 0.32 $\pm$ 0.18 & 235.69 $\pm$ 415.15 \\
 & LAPEFT+T5Chem & 0.53 $\pm$ 0.06 & 249.48 $\pm$ 192.73 \\
\hline
\nopagebreak
\multirow{7}{*}{BH-3} & PLLM$\phi$+T5 (ours) & 0.47 $\pm$ 0.13 & 6.69 $\pm$ 1.28 \\
 & PLLM$\phi$+T5Chem (ours) & 0.50 $\pm$ 0.10 & 8.34 $\pm$ 1.58 \\
 & GP+DRFP & 0.38 $\pm$ 0.30 & 5.72 $\pm$ 8.59 \\
 & Bochem.+T5 & 0.16 $\pm$ 0.14 & 22.18 $\pm$ 47.62 \\
 & Bochem.+T5Chem & 0.41 $\pm$ 0.13 & 50.64 $\pm$ 99.46 \\
 & LAPEFT+T5 & 0.27 $\pm$ 0.10 & 71.06 $\pm$ 53.91 \\
 & LAPEFT+T5Chem & 0.40 $\pm$ 0.07 & 350.63 $\pm$ 528.16 \\
\hline
\nopagebreak
\multirow{7}{*}{BH-4} & PLLM$\phi$+T5 (ours) & 0.57 $\pm$ 0.08 & 7.86 $\pm$ 1.15 \\
 & PLLM$\phi$+T5Chem (ours) & 0.59 $\pm$ 0.12 & 8.72 $\pm$ 1.47 \\
 & GP+DRFP & 0.48 $\pm$ 0.29 & 4.13 $\pm$ 0.57 \\
 & Bochem.+T5 & 0.46 $\pm$ 0.15 & 39.56 $\pm$ 43.63 \\
 & Bochem.+T5Chem & 0.46 $\pm$ 0.13 & 41.01 $\pm$ 54.16 \\
 & LAPEFT+T5 & 0.32 $\pm$ 0.09 & 80.44 $\pm$ 89.41 \\
 & LAPEFT+T5Chem & 0.40 $\pm$ 0.08 & 581.34 $\pm$ 2146.82 \\
\hline
\nopagebreak
\multirow{7}{*}{BH-5} & PLLM$\phi$+T5 (ours) & 0.50 $\pm$ 0.10 & 8.48 $\pm$ 1.42 \\
 & PLLM$\phi$+T5Chem (ours) & 0.53 $\pm$ 0.08 & 9.08 $\pm$ 1.48 \\
 & GP+DRFP & 0.42 $\pm$ 0.28 & 4.31 $\pm$ 0.59 \\
 & Bochem.+T5 & 0.33 $\pm$ 0.17 & 49.31 $\pm$ 62.92 \\
 & Bochem.+T5Chem & 0.43 $\pm$ 0.10 & 57.96 $\pm$ 51.71 \\
 & LAPEFT+T5 & 0.37 $\pm$ 0.11 & 63.98 $\pm$ 44.88 \\
 & LAPEFT+T5Chem & 0.42 $\pm$ 0.10 & 391.56 $\pm$ 543.81 \\
\hline
\nopagebreak
\multirow{7}{*}{C2 yield} & PLLM$\phi$+T5 (ours) & 0.44 $\pm$ 0.09 & 4.83 $\pm$ 2.30 \\
 & PLLM$\phi$+T5Chem (ours) & 0.42 $\pm$ 0.12 & 4.56 $\pm$ 1.73 \\
 & GP+Num.Params & 0.16 $\pm$ 0.09 & 154792.41 $\pm$ 578478.07 \\
 & Bochem.+T5 & 0.41 $\pm$ 0.11 & 0.99 $\pm$ 0.94 \\
 & Bochem.+T5Chem & 0.42 $\pm$ 0.14 & 5.57 $\pm$ 18.91 \\
 & LAPEFT+T5 & 0.29 $\pm$ 0.09 & 180.61 $\pm$ 220.08 \\
 & LAPEFT+T5Chem & 0.24 $\pm$ 0.09 & 1196.45 $\pm$ 1636.29 \\
\hline
\nopagebreak
\multirow{7}{*}{HPLC} & PLLM$\phi$+T5 (ours) & -0.13 $\pm$ 0.07 & 20.13 $\pm$ 5.14 \\
 & PLLM$\phi$+T5Chem (ours) & -0.21 $\pm$ 0.07 & 20.18 $\pm$ 5.66 \\
 & GP+Num.Params & -0.04 $\pm$ 0.05 & 1444.58 $\pm$ 6420.47 \\
 & Bochem.+T5 & -0.03 $\pm$ 0.03 & 18.57 $\pm$ 14.46 \\
 & Bochem.+T5Chem & -0.02 $\pm$ 0.03 & 49.10 $\pm$ 66.46 \\
 & LAPEFT+T5 & -0.22 $\pm$ 0.14 & 229.99 $\pm$ 255.04 \\
 & LAPEFT+T5Chem & -0.18 $\pm$ 0.09 & 2427.96 $\pm$ 8532.05 \\
\hline
\nopagebreak
\multirow{7}{*}{KINASE} & PLLM$\phi$+T5 (ours) & 0.35 $\pm$ 0.06 & 14.55 $\pm$ 2.16 \\
 & PLLM$\phi$+T5Chem (ours) & 0.41 $\pm$ 0.05 & 12.82 $\pm$ 2.09 \\
 & GP+FP & -0.02 $\pm$ 0.03 & 1.41 $\pm$ 0.11 \\
 & Bochem.+T5 & 0.02 $\pm$ 0.11 & 3.44 $\pm$ 6.44 \\
 & Bochem.+T5Chem & -0.02 $\pm$ 0.02 & 1.67 $\pm$ 0.11 \\
 & LAPEFT+T5 & 0.33 $\pm$ 0.07 & 12.42 $\pm$ 3.44 \\
 & LAPEFT+T5Chem & 0.45 $\pm$ 0.04 & 21.59 $\pm$ 5.58 \\
\hline
\nopagebreak
\multirow{7}{*}{OER} & PLLM$\phi$+T5 (ours) & 0.43 $\pm$ 0.15 & 1.95 $\pm$ 1.61 \\
 & PLLM$\phi$+T5Chem (ours) & 0.42 $\pm$ 0.15 & 0.88 $\pm$ 1.31 \\
 & GP+Num.Params & 0.59 $\pm$ 0.04 & 189.97 $\pm$ 218.23 \\
 & Bochem.+T5 & 0.44 $\pm$ 0.09 & 17.26 $\pm$ 23.03 \\
 & Bochem.+T5Chem & 0.44 $\pm$ 0.11 & 154.70 $\pm$ 206.00 \\
 & LAPEFT+T5 & 0.47 $\pm$ 0.08 & 27.27 $\pm$ 34.99 \\
 & LAPEFT+T5Chem & 0.47 $\pm$ 0.07 & 174.81 $\pm$ 286.01 \\
\hline
\nopagebreak
\multirow{7}{*}{PCE} & PLLM$\phi$+T5 (ours) & 0.25 $\pm$ 0.12 & 16.12 $\pm$ 4.78 \\
 & PLLM$\phi$+T5Chem (ours) & 0.27 $\pm$ 0.11 & 15.43 $\pm$ 4.42 \\
 & GP+FP & -0.02 $\pm$ 0.02 & 2.52 $\pm$ 0.08 \\
 & Bochem.+T5 & 0.10 $\pm$ 0.15 & 4.51 $\pm$ 6.22 \\
 & Bochem.+T5Chem & -0.02 $\pm$ 0.03 & 2.57 $\pm$ 0.15 \\
 & LAPEFT+T5 & 0.16 $\pm$ 0.15 & 29.69 $\pm$ 46.25 \\
 & LAPEFT+T5Chem & 0.26 $\pm$ 0.13 & 41.33 $\pm$ 21.10 \\
\hline
\nopagebreak
\multirow{7}{*}{PHOTOSWITCH} & PLLM$\phi$+T5 (ours) & 0.59 $\pm$ 0.12 & 11.53 $\pm$ 2.13 \\
 & PLLM$\phi$+T5Chem (ours) & 0.65 $\pm$ 0.10 & 9.86 $\pm$ 1.62 \\
 & GP+FP & 0.57 $\pm$ 0.26 & 5.13 $\pm$ 0.58 \\
 & Bochem.+T5 & 0.58 $\pm$ 0.07 & 41238535.40 $\pm$ 32024292.47 \\
 & Bochem.+T5Chem & 0.46 $\pm$ 0.28 & 151181279.47 $\pm$ 202238451.62 \\
 & LAPEFT+T5 & 0.52 $\pm$ 0.08 & 7.78 $\pm$ 1.20 \\
 & LAPEFT+T5Chem & 0.64 $\pm$ 0.07 & 18.50 $\pm$ 11.90 \\
\hline
\nopagebreak
\multirow{7}{*}{REDOX-MER} & PLLM$\phi$+T5 (ours) & 0.86 $\pm$ 0.02 & 0.48 $\pm$ 0.47 \\
 & PLLM$\phi$+T5Chem (ours) & 0.89 $\pm$ 0.02 & -0.02 $\pm$ 0.35 \\
 & GP+FP & 0.73 $\pm$ 0.25 & -0.51 $\pm$ 0.44 \\
 & Bochem.+T5 & 0.85 $\pm$ 0.02 & 19.72 $\pm$ 16.26 \\
 & Bochem.+T5Chem & 0.85 $\pm$ 0.20 & 47.01 $\pm$ 40.57 \\
 & LAPEFT+T5 & 0.73 $\pm$ 0.06 & 6.45 $\pm$ 2.53 \\
 & LAPEFT+T5Chem & 0.86 $\pm$ 0.03 & 16.55 $\pm$ 6.20 \\
\hline
\nopagebreak
\multirow{7}{*}{SOLVATION} & PLLM$\phi$+T5 (ours) & 0.74 $\pm$ 0.03 & 2.84 $\pm$ 0.98 \\
 & PLLM$\phi$+T5Chem (ours) & 0.73 $\pm$ 0.04 & 2.86 $\pm$ 0.97 \\
 & GP+FP & 0.77 $\pm$ 0.03 & -0.36 $\pm$ 0.47 \\
 & Bochem.+T5 & 0.79 $\pm$ 0.03 & 77.92 $\pm$ 58.50 \\
 & Bochem.+T5Chem & 0.69 $\pm$ 0.30 & 174.24 $\pm$ 192.44 \\
 & LAPEFT+T5 & 0.71 $\pm$ 0.05 & 14.28 $\pm$ 20.01 \\
 & LAPEFT+T5Chem & 0.79 $\pm$ 0.02 & 73.66 $\pm$ 38.83 \\
\hline
\nopagebreak
\multirow{7}{*}{SUZUKI-MIYAURA} & PLLM$\phi$+T5 (ours) & 0.10 $\pm$ 0.13 & 8.16 $\pm$ 1.45 \\
 & PLLM$\phi$+T5Chem (ours) & 0.12 $\pm$ 0.13 & 9.65 $\pm$ 2.37 \\
 & GP+DRFP & 0.29 $\pm$ 0.15 & 1.20 $\pm$ 2.57 \\
 & Bochem.+T5 & 0.07 $\pm$ 0.08 & 8.76 $\pm$ 28.10 \\
 & Bochem.+T5Chem & 0.12 $\pm$ 0.10 & 1.11 $\pm$ 3.25 \\
 & LAPEFT+T5 & -0.10 $\pm$ 0.18 & 107.12 $\pm$ 105.69 \\
 & LAPEFT+T5Chem & 0.18 $\pm$ 0.09 & 231.35 $\pm$ 181.75 \\
\hline
\nopagebreak
\multirow{7}{*}{VAPDIFF} & PLLM$\phi$+T5 (ours) & -0.01 $\pm$ 0.06 & 12.40 $\pm$ 2.73 \\
 & PLLM$\phi$+T5Chem (ours) & -0.05 $\pm$ 0.09 & 14.25 $\pm$ 3.41 \\
 & GP+Num.Params & 0.12 $\pm$ 0.06 & 608876.27 $\pm$ 906151.79 \\
 & Bochem.+T5 & 0.07 $\pm$ 0.06 & 1.69 $\pm$ 0.28 \\
 & Bochem.+T5Chem & 0.08 $\pm$ 0.07 & 1.72 $\pm$ 0.46 \\
 & LAPEFT+T5 & -0.17 $\pm$ 0.11 & 288.12 $\pm$ 284.60 \\
 & LAPEFT+T5Chem & -0.04 $\pm$ 0.08 & 164.06 $\pm$ 142.57 \\
\hline
\caption{Predictive and uncertainty estimates for all benchmark datasets and methods. Each model is trained on 60 points and evaluated on the remaining data, emulating a 10+50 BO iteration setup. This fixed train/validation split ensures fair comparison by avoiding divergence in design sets during BO due to different selection of candidate points during optimization, even when starting with the same initial points. We run 20 repeats and report mean and standar deviation values.}
\end{longtable}

\begin{table}[htbp]  
\small
\centering
\begin{tabular}{lllc}
\hline
Arch. & Pool & Model & Quant. 95 [cnt] \\
\hline
\multirow{9}{*}{Enc} 
    & \multirow{3}{*}{CLS} & MXBAI        & 2.70 $\pm$ 2.15 \\
    &                      & ModernBERT   & 2.40 $\pm$ 2.09 \\
    &                      & UAE          & 2.40 $\pm$ 2.11 \\

    & \multirow{3}{*}{Avg} & MXBAI        & 2.30 $\pm$ 2.00 \\
    &                      & ModernBERT   & 2.30 $\pm$ 1.72 \\
    &                      & UAE          & 2.25 $\pm$ 1.86 \\

    & \multirow{3}{*}{Last} & MXBAI       & 1.80 $\pm$ 1.82 \\
    &                       & ModernBERT  & 0.25 $\pm$ 0.55 \\
    &                       & UAE         & 2.50 $\pm$ 2.35 \\
\hline
\multirow{4}{*}{Dec} 
    & \multirow{2}{*}{Avg}  & LLama3-8B   & 0.75 $\pm$ 1.48 \\
    &                       & Qwen2-7B    & 1.05 $\pm$ 2.28 \\
    & \multirow{2}{*}{Last} & LLama3-8B   & 1.55 $\pm$ 1.64 \\
    &                       & Qwen2-7B    & 1.65 $\pm$ 2.54 \\
\hline
\multirow{4}{*}{Enc-Dec} 
    & \multirow{2}{*}{Avg}  & T5          & 1.10 $\pm$ 1.41 \\
    &                       & T5Chem      & 1.90 $\pm$ 2.05 \\
    & \multirow{2}{*}{Last} & T5          & 1.65 $\pm$ 2.48 \\
    &                       & T5Chem      & 1.55 $\pm$ 2.91 \\
\hline
\end{tabular}
\caption{Tokenization influence to sampling from the 5th percentile. }
\label{tab:tokenization}
\end{table}

\begin{table}[htbp]
\centering
\begin{tabular}{llllc}
\hline
Architecture & Target layers & Target ratio & Deep Kernel + Model & Quantile 95 [cnt] \\
\hline
\multirow{6}{*}{Chem-related} 
    & \multirow{3}{*}{Top} & 0.1  & \multirow{6}{*}{PLLM$\phi$+T5Chem-SMILES} & 6.60 $\pm$ 1.90 \\
    &                      & 0.25 &  & 6.30 $\pm$ 3.47 \\
    &                      & 0.5  &  & 8.20 $\pm$ 2.94 \\
    & \multirow{3}{*}{Bottom} & 0.1  &  & 7.50 $\pm$ 2.51 \\
    &                      & 0.25 &  & 7.40 $\pm$ 2.32 \\
    &                      & 0.5  &  & 6.60 $\pm$ 3.13 \\
\hline
\multirow{6}{*}{Encoder} 
    & \multirow{3}{*}{Top} & 0.1  & \multirow{6}{*}{PLLM$\phi$+ModernBERT} & 5.50 $\pm$ 2.17 \\
    &                      & 0.25 &  & 5.60 $\pm$ 2.37 \\
    &                      & 0.5  &  & 5.70 $\pm$ 2.83 \\
    & \multirow{3}{*}{Bottom} & 0.1  &  & 4.20 $\pm$ 2.44 \\
    &                      & 0.25 &  & 4.50 $\pm$ 3.21 \\
    &                      & 0.5  &  & 4.30 $\pm$ 2.11 \\
\hline
\multirow{7}{*}{Enc-Dec} 
    & \multirow{3}{*}{Top} & 0.1  & \multirow{7}{*}{PLLM$\phi$+T5} & 4.10 $\pm$ 1.91 \\
    &                      & 0.25 &  & 5.40 $\pm$ 2.99 \\
    &                      & 0.5  &  & 3.90 $\pm$ 3.25 \\
    & \multirow{4}{*}{Bottom} & 0.1  &  & 4.90 $\pm$ 2.08 \\
    &                      & 0.25 &  & 5.40 $\pm$ 2.55 \\
    &                      & 0.5  &  & 4.90 $\pm$ 2.18 \\
\hline
\multirow{2}{*}{Decoder} 
    & \multirow{2}{*}{Top} & 0.1  & \multirow{2}{*}{PLLM$\phi$+Qwen2-7B} & 4.10 $\pm$ 2.28 \\
    &                      & 0.25 &  & 4.90 $\pm$ 2.88 \\
\hline
\end{tabular}
\caption{GP-LLM finetuning per LLM types and LoRA layers}
\label{tab:peft_layer_targets}
\end{table}

\end{document}